\PassOptionsToPackage{table}{xcolor}
\documentclass[final,5p,times,twocolumn]{elsarticle}
\usepackage[T1]{fontenc}
\usepackage{lmodern}

\usepackage[utf8]{inputenc}
\usepackage{amsmath,amssymb,booktabs,tabularx,longtable,graphicx}
\usepackage{setspace,xcolor,caption,microtype}
\usepackage[colorlinks=true,linkcolor=black,citecolor=black,urlcolor=black,breaklinks=true]{hyperref}
\journal{}
\biboptions{sort&compress}

\NewDocumentEnvironment{widefigure}{+b}{\begin{figure*}[!t]\centering #1\end{figure*}}{}
\NewDocumentEnvironment{widetable}{+b}{\begin{table*}[!t]\centering #1\end{table*}}{}
\newcommand{\ind}{\mathbf{1}}

\definecolor{OurBlue}{RGB}{232,243,252}
\newcounter{algorithm}
\newenvironment{paperalgorithm}[1]{\refstepcounter{algorithm}\noindent\begin{minipage}{\linewidth}\hrule height .7pt\vspace{5pt}\textbf{Algorithm~\thealgorithm. #1}\par\vspace{4pt}\hrule height .35pt\vspace{6pt}}{\par\vspace{6pt}\hrule height .7pt\end{minipage}}


\newcommand{\RowsEBFDAblation}{%
FMP reader & 95.01 & 90.88 & 91.33 & 89.47 \\
FMP + guarded geometry & \underline{98.08} & \underline{93.54} & 91.33 & 89.47 \\
Direct DFD: plain set & 95.01 & 90.88 & \underline{93.41} & 91.78 \\
Direct DFD: evidence edit & 95.42 & 92.41 & 92.64 & \underline{91.83} \\
Hybrid: plain set & 98.02 & \underline{93.54} & 93.38 & 91.75 \\
Bound DFD: initial readout & \textbf{98.19} & \textbf{94.52} & 92.63 & 91.79 \\
\rowcolor{OurBlue}SGDB: certified readout & \textbf{98.19} & \textbf{94.52} & \textbf{93.56} & \textbf{92.76} \\
}

\newcommand{\RowsReadoutAblation}{%
Initial bound readout & \textbf{98.19} & \textbf{94.52} & 92.63 & 91.79 \\
Uniform mask readout & 96.36 & 93.54 & \textbf{93.58} & \textbf{92.86} \\
Quality selection only & \underline{96.62} & \underline{93.82} & 93.54 & \underline{92.76} \\
\rowcolor{OurBlue}SGDB: complete support & \textbf{98.19} & \textbf{94.52} & \underline{93.56} & \underline{92.76} \\
}

\newcommand{\RowsBindingPolicies}{%
\rowcolor{OurBlue}SGDB partial binding & \textbf{98.19} & \textbf{94.52} & \underline{93.56} & \underline{92.76} \\
Independent nearest witness & \textbf{98.19} & \textbf{94.52} & \underline{93.56} & \underline{92.76} \\
Without class compatibility & \textbf{98.19} & \textbf{94.52} & 92.60 & 91.38 \\
Without distance gate & \textbf{98.19} & \textbf{94.52} & \underline{93.56} & \underline{92.76} \\
Discard unbound identities & \textbf{98.19} & \textbf{94.52} & \underline{93.56} & \underline{92.76} \\
Tighter radius (0.04) & \underline{97.98} & \textbf{94.52} & \textbf{93.97} & \textbf{93.11} \\
}

\newcommand{\RowsExplanation}{%
All executed rules & 7.121 & \textbf{100.00} & \textbf{0} \\
All positive-score rules & 1.994 & \textbf{100.00} & \textbf{0} \\
Maximum-score rule & 1.000 & \underline{99.70} & \underline{17} \\
\rowcolor{OurBlue}Compact sufficient (SGDB) & 1.006 & \textbf{100.00} & \textbf{0} \\
}

\begin{document}
\hypersetup{hidelinks,pdftitle={SGDB: Semantic–Geometric Decoupling and Evidence Binding for Logical Anomaly Detection},pdfauthor={Chenrui Ji; Jiafan Lu; Zilu Luo; Feng Wan; Qiuhui Chen; Weimin Zhong}}
\begin{frontmatter}
\title{SGDB: Semantic--Geometric Decoupling\\and Evidence Binding for Logical Anomaly Detection}
\author[ecust]{Chenrui Ji}
\author[ecust]{Jiafan Lu}
\author[ecust]{Zilu Luo}
\author[ecust]{Feng Wan}
\author[ecust]{Qiuhui Chen\corref{corresponding}}
\ead{chenqh@ecust.edu.cn}
\author[ecust]{Weimin Zhong\corref{corresponding}}
\ead{wmzhong@ecust.edu.cn}
\cortext[corresponding]{Corresponding authors: Qiuhui Chen and Weimin Zhong.}
\address[ecust]{School of Information Science and Engineering, East China University of Science and Technology, Shanghai 200237, China}
\begin{abstract}
Logical anomaly detection requires recognizing scene components and measuring whether their configuration satisfies task constraints. Correct identities can coexist with inaccurate geometry; clean mask fragments can yield misleading lengths. We propose Semantic--Geometric Decoupling and Evidence Binding (SGDB). A semantic decoder learns object identities from component-annotated normal sources and factual edits; a separate reader supplies final measurements. Partial one-to-one association binds geometric witnesses to identities while retaining unmatched objects in the count. A support-certified reader selects mask-derived axes using line quality and complete component ownership. Explicit task requirements and normal-data references define executable checks, with decision thresholds fixed by independent normal calibration. Execution-grounded explanation traces rule contributions to source objects and measurements, extracting compact sufficient alarm certificates. A frozen language model expresses selected outcomes as cited claims, with numeric evidence retained by the executor. Reports preserve observed deviations and unavailable evidence. SGDB achieves 98.19\% AUROC and 94.52\% fixed-threshold F1 on five MVTec LOCO logical categories, and 93.56\% AUROC and 92.76\% F1 across all 50 VID-AD tasks. Frozen-output controls evaluate geometric readout and explanatory sufficiency and compactness.
\end{abstract}
\end{frontmatter}
\thispagestyle{plain}
\raggedbottom
\section{Introduction}
Logical anomaly detection requires recognizing scene components and measuring whether their configuration satisfies task requirements. Two screw bags can contain identical component counts yet differ in geometric validity; a tray can contain the required utensils in the wrong order. Recognition establishes identities and counts, while measurement supplies lengths, positions and relations. MVTec LOCO AD distinguishes these logical violations from structural defects \cite{loco}, and VID-AD tests inspection under capture distractions \cite{vid}.

\begin{figure}[!t]
\centering\includegraphics[width=\columnwidth]{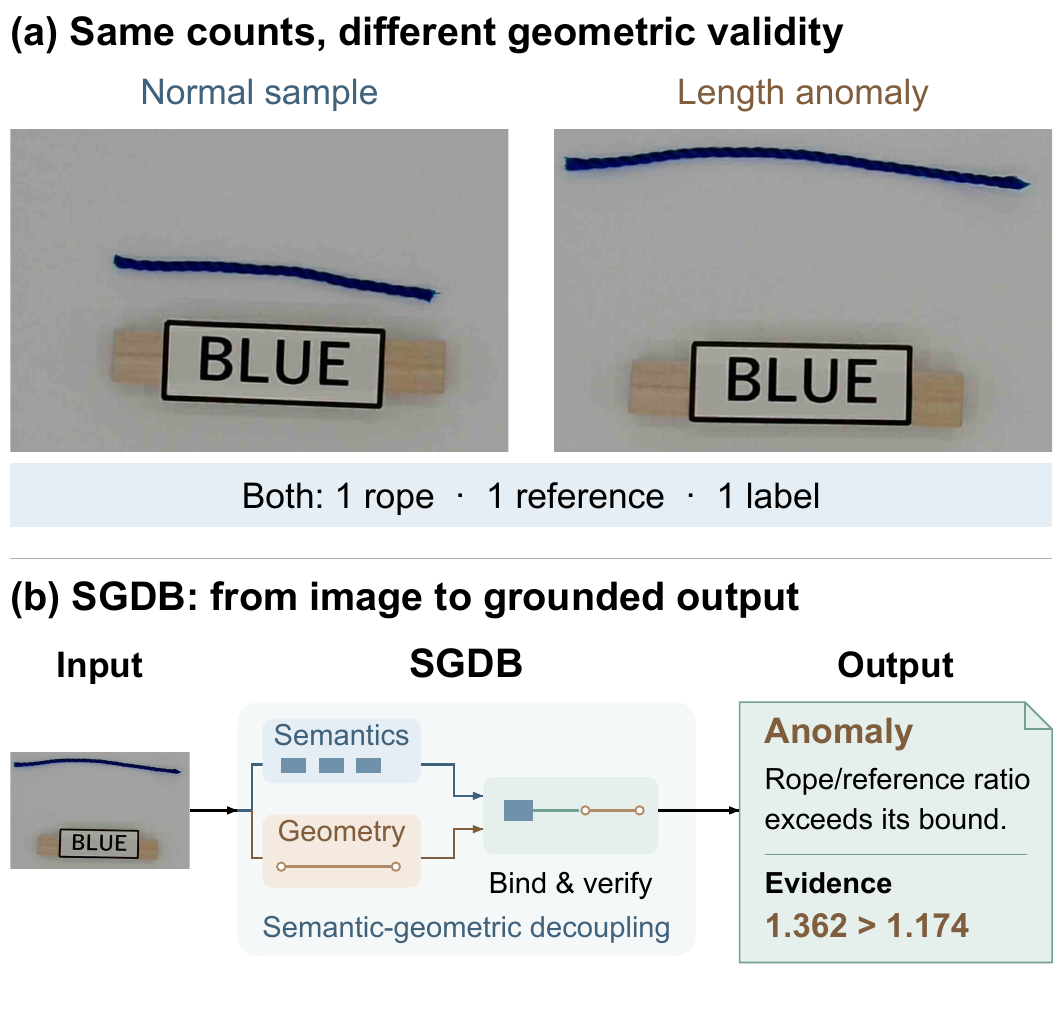}
\captionsetup{font={footnotesize,stretch=1.0}}
\caption{Motivation for explicit geometric verification beyond appearance/composition scoring \cite{patchcore,salad}. (a) The two scenes have identical component counts. (b) SGDB maps an image to a decision, explanation and evidence through semantic--geometric decoupling, binding and verification. The recorded rope/reference ratio 1.362 exceeds the normal upper bound 1.174. Images and measurements are from frozen outputs.}
\label{fig:motivation}
\end{figure}

Recent component-aware methods model component regions, object composition or reference matching \cite{psad,salad,samlad,objectcore}. Language-based approaches express inspection requirements through descriptions, programs and constraint subqueries \cite{logicad,logicode,logicls}. Figure~\ref{fig:motivation} illustrates why correct object identities and counts do not ensure geometric validity. Each geometric operand must be suitable for its constraint and attached to the intended object. A correctly recognized screw can have inaccurate regressed endpoints. A mask-derived length requires an identity to be checked against its object-specific bound. However, assigning a clean fragment of a merged region to an object can produce a plausible but incomplete axis. Readout must therefore check the instance support as well as the quality of the fitted line.

We study logical inspection configured with normal images, component annotations and explicit task requirements. Component targets supervise visual readers; normal observations establish measurement ranges and score references; task requirements specify the discrete constraints to execute. We propose \emph{Semantic--Geometric Decoupling and Evidence Binding} (SGDB), which separates semantic recognition from final geometric readout and reconnects their outputs through explicit evidence binding. Component-aware inspection \cite{comad,csad,salad,logico} provides useful object and region representations; binding establishes which measured operand belongs to each decoded identity.

SGDB implements this separation above a shared visual frontend. A learned object-set decoder reads frozen features with candidate and spatial queries. A witness reader supplies initial geometric operands, including native-coordinate mask readout for screw axes. Same-class partial one-to-one association attaches compatible witnesses to decoded objects using center proximity. A subsequent support-certified readout replaces a length axis only when the candidate improves support quality and preserves a complete, distinct component for its observed instance. Each fact retains class, confidence, associated geometry, measurement availability and source indices. Unmatched identities still contribute to the count, with their geometric operands marked unavailable.

Bound facts support both executable verification and an explanation of the resulting decision. A rule deviation alone need not trigger an alarm: normal-reference normalization, aggregation and the task threshold determine its effect. SGDB therefore records each executed operand, requirement and score contribution together with its source links. It extracts a compact sufficient alarm certificate and uses the same trace to express the decision, supporting measurements and unavailable evidence.

The selected trace also conditions a frozen language model, whose claims retain executor-owned evidence citations and numeric records.

SGDB connects three complementary mechanisms through this evidence representation:

\begin{enumerate}
\item Semantic--geometric fact construction assigns object identification and counting to a learned semantic decoder and final geometric operands to a separate witness readout.
\item Partial evidence binding and support-certified readout join identities to compatible measurements, preserve unmatched identities and verify component ownership before updating geometry.
\item Execution-grounded explanation extracts compact sufficient alarm certificates and traces their predicates to object identities, measured operands and evidence availability.
\end{enumerate}
Evaluation spans all five LOCO logical categories and all 50 VID tasks. Shared-supervision controls examine fact construction and readout; frozen-trace comparisons examine explanatory sufficiency and compactness.

\section{Related work}
\subsection{Appearance models and industrial benchmarks}
Normal-appearance modeling is a central approach to industrial anomaly detection. MVTec AD provides defect images with pixel annotations \cite{mvtecad}, while VisA broadens the range of objects and multi-instance scenes \cite{visa}. PaDiM models patch embeddings with multivariate Gaussian distributions \cite{padim}, and PatchCore stores a representative memory of normal patch features \cite{patchcore}. Reverse Distillation reconstructs teacher features from a compact one-class embedding \cite{rd}; EfficientAD combines efficient feature prediction with a global autoencoder \cite{efficientad}. DRAEM instead learns reconstruction and discrimination jointly using simulated anomalies \cite{draem}. These approaches establish useful appearance representations. Logical inspection additionally requires the identities, quantities and relationships of otherwise plausible components to be evaluated against the task configuration \cite{loco}.

% Queue the overview in late related work so the method figures occupy different pages.
\begin{figure*}[!t]
\centering\includegraphics[width=\textwidth]{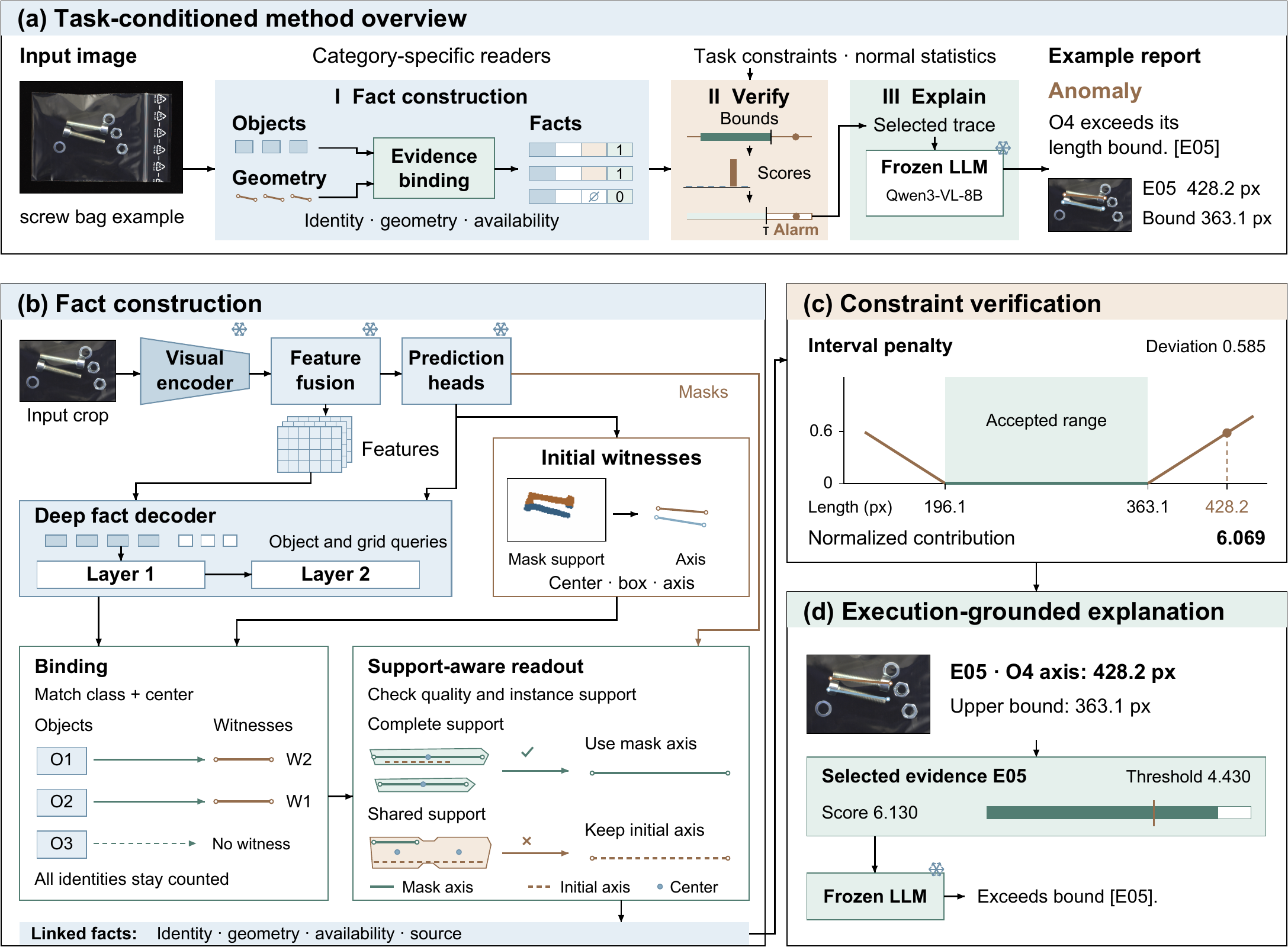}
\captionsetup{font={footnotesize,stretch=1.0}}
\caption{SGDB framework. (a) Bound facts are checked against task constraints, aggregated and compared with the alarm threshold. (b) Objects bind to geometric witnesses by class and position; unmatched identities remain counted. Quality and instance-support checks select the mask axis or retain the initial axis, shown by solid green and dashed brown lines. (c) The interval penalty is zero within the accepted range and increases linearly outside it; the marked point is the recorded O4 length. (d) Selected evidence E05 supports the frozen LLM's source-attributed explanation; numeric evidence is presented by the executor. Snowflakes denote the frozen visual frontend and language model.}
\label{fig:framework}
\end{figure*}

\subsection{Component and relationship representations}
Component-level methods address this requirement through segmentation and composition. ComAD extracts metrological properties from component regions \cite{comad}. PSAD uses a few annotated images to learn part segmentation and combines histogram, composition and patch memories \cite{psad}; CSAD obtains component pseudo-labels and combines patch histograms with local--global modeling \cite{csad}. SALAD learns the distribution of composition maps through synthetic anomalies \cite{salad}, whereas LogiCo combines component reconstruction with segmentation-map discrimination \cite{logico}. PUAD complements reconstruction with an out-of-distribution model of spatially pooled features \cite{puad}. Recent work in Neurocomputing also addresses global context through pyramid fusion and state-space enhancement \cite{neuro_mamba}, and represents quantity and spatial relationships with separate memories \cite{neuro_relation}.

Object matching provides a more explicit representation of composition. SAM-LAD matches segmented objects between query and reference images \cite{samlad}. ObjectCore represents an image as a collection of object features and detects few-shot logical anomalies through bipartite matching to a support image \cite{objectcore}. SGDB uses association within the same image: decoded semantic identities are linked to measurements produced by a separate visual reader. This yields object-specific operands for subsequent constraint execution, including counted identities whose geometry is unavailable.

Synthetic logical examples also provide supervision when real violations are scarce. ComGEN generates anomalies through component-aware compositional editing \cite{logicalgen}. Chimera training constructs counterfactual operands at the feature level to train a neural rule evaluator \cite{chimera}. SGDB uses factual image edits with updated component targets to train semantic decoding; task constraints are evaluated by a fixed executor after evidence association.

\subsection{Object queries and geometric evidence}
Attention supplies a general mechanism for reading visual features \cite{attention}. DETR formulates detection as object-set prediction with bipartite training assignment \cite{detr}. Deformable DETR conditions attention on a small set of locations around reference points \cite{deformable}, while DAB-DETR expresses queries using dynamically updated anchor boxes \cite{dab}. These designs show how spatial hypotheses can guide semantic feature extraction. SGDB combines candidate-initialized and grid queries, then separates semantic decoding from the selection of a final geometric operand. Its deployment association links two prediction sets, rather than predictions and training targets.

When constraints depend on physical image geometry, the coordinate system of a measurement also matters. Anisotropic resizing can alter Euclidean lengths and principal directions. SGDB reads screw axes through instance-bound PCA in native coordinates \cite{pca}, with support and elongation checks that retain the existing axis when the mask readout is unsuitable.

\subsection{Language, executable reasoning and reports}
Vision--language approaches connect visual normality to textual descriptions. WinCLIP uses prompt ensembles and window-level features \cite{winclip}; AnomalyCLIP learns object-agnostic normality prompts \cite{anomalyclip}. AnomalyGPT supports anomaly localization and interactive descriptions \cite{anomalygpt}. For logical inspection, LogicAD combines extracted descriptions with formal reasoning \cite{logicad}, LogiCode generates executable inspection programs \cite{logicode}, and LogiCls decomposes constraints into subqueries for instruction-tuned violation classification \cite{logicls}. VID-AD examines description-based logical inspection under visual distraction \cite{vid}.

Executable intermediate representations are also studied beyond anomaly detection. VISPROG composes visual modules into programs with inspectable intermediate outputs \cite{visprog}. DeepProbLog integrates neural predicates with probabilistic logic programs \cite{deepproblog}, and Logic Tensor Networks ground logical expressions in differentiable computations \cite{ltn}. NeSyLAD applies learned segmentation and symbolic rule extraction to industrial logical anomalies \cite{nesylad}. SGDB evaluates explicit discrete requirements and continuous intervals fitted from normal data. Frozen OCR observations remain inputs for rope-label tasks; Qwen3-VL is evaluated as a separate comparison deployment \cite{qwen25,qwen3}.

Sufficient explanations identify compact evidence supporting a prediction. Sufficient-subset training learns concise explanatory outputs \cite{sst}, while sufficiency and necessity characterize different aspects of explanatory importance \cite{suffnecessary}. SGDB constructs its certificate over executed normalized rule scores. The certificate is connected to the semantic and geometric sources already established by evidence binding, and the full trace retains the other predicates and unavailable operands.

% Let the method follow related work without an empty remainder column.
\section{Semantic--geometric decoupling and evidence binding}
\subsection{Task setup and fact representation}
Given image $x\in\mathbb R^{H\times W\times3}$ and task $t$, SGDB returns an anomaly score $s_t(x)$, decision $\hat y_t(x)$ and supporting report. \mbox{Figure~\ref{fig:framework}} organizes the method into fact decoding, constraint verification and explanation. Its central representation is
\begin{equation}
 f_i=(\hat c_i,p_i,g_i,a_i),
\end{equation}
where $\hat c_i$ and $p_i$ are a decoded class and confidence, $g_i$ is associated geometric evidence, and $a_i$ indicates measurement availability. Task rules operate on these assembled facts.

Component targets on normal fitting/development images supervise the visual readers and their selection. Explicit task requirements configure the executor; normal measurements supply continuous bounds and score references. At inspection, the input is an image and its task identifier. The readers predict object identities, masks and geometric witnesses, which are assembled into the facts above.

\subsection{Semantic object decoding}
The visual frontend comprises an ImageNet-initialized WideResNet50-2 encoder \cite{imagenet,wrn}, multiscale feature fusion and center, box, semantic-mask and endpoint heads. Center localization follows the object-as-point formulation \cite{objects}. Component targets from normal sources supervise the frontend, which is frozen during semantic-decoder training.

\begin{figure}[!t]
\centering\includegraphics[width=\columnwidth]{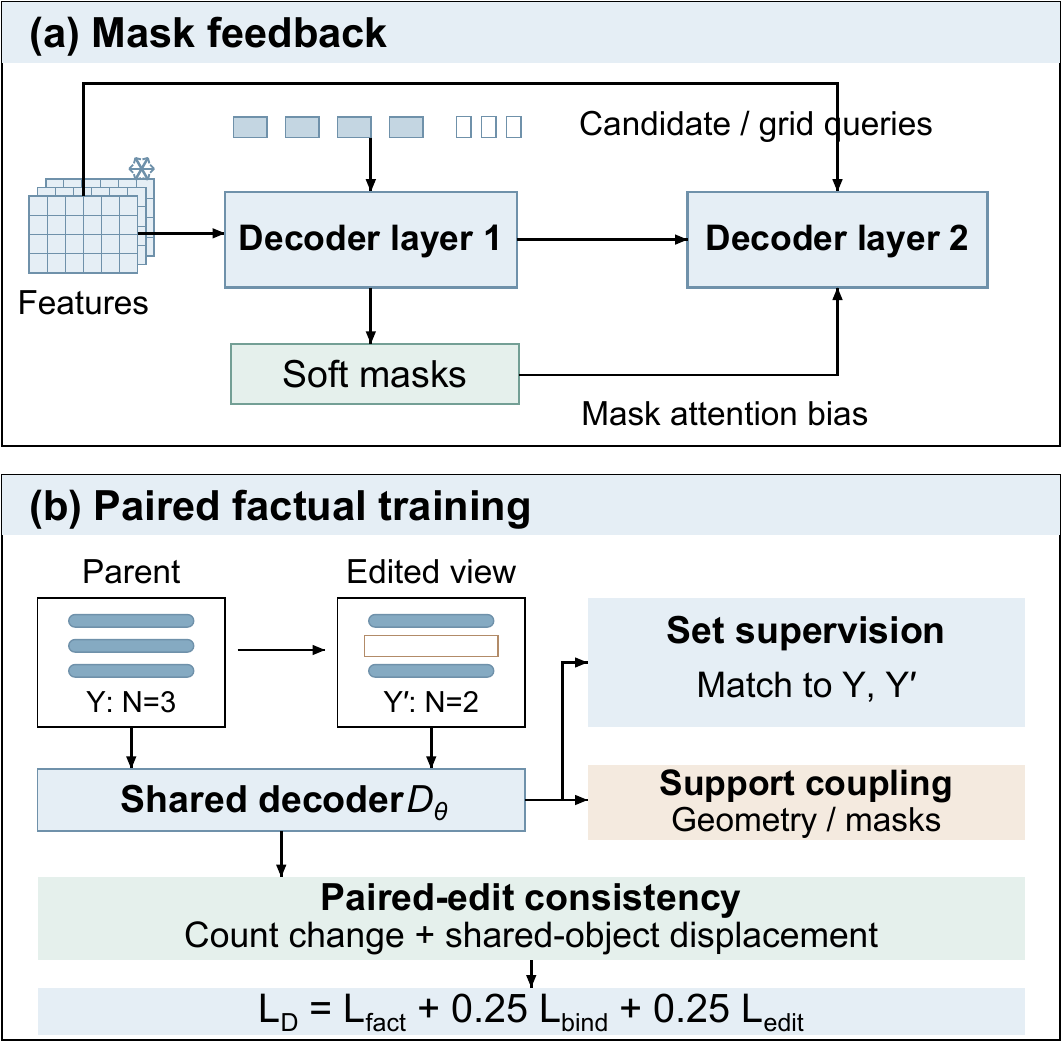}
\captionsetup{font={footnotesize,stretch=1.0}}
\caption{Fact-decoder training. (a) First-layer soft masks bias second-layer attention over frozen visual features. (b) A normal-source parent and its edit share decoder parameters. Updated targets supervise object sets; support coupling aligns geometry with masks, and edit consistency supervises count changes and shared-object displacement. The removal example is schematic.}
\label{fig:decodertraining}
\end{figure}

The decoder reads dense features $F(x)$ using candidate-initialized queries and spatial queries:
\begin{equation}
\begin{aligned}
 O(x)&=D_\theta(F(x),Q_{\rm cand},Q_{\rm grid})\\
     &=\{(\hat c_i,p_i,\hat{\mathbf u}_i)\}_{i=1}^{n}.
\end{aligned}
\end{equation}
Candidate queries encode the frontend's class, confidence and geometric hypothesis together with sampled visual features. Spatial queries cover a fixed grid. Two Transformer layers update the queries; first-layer soft support masks contribute a spatial attention bias to the second layer. Figure~\ref{fig:decodertraining} details this training mechanism. A normal-source parent and its factual edit share decoder parameters. Hungarian assignment matches each prediction set to its updated component targets for class, center, box, axis and mask supervision, denoted by $\mathcal L_{\rm fact}$. Support coupling $\mathcal L_{\rm bind}$ aligns predicted geometry with mask support. Paired-edit consistency $\mathcal L_{\rm edit}$ supervises count changes and shared-object displacements:
\begin{equation}
\mathcal L_D=\mathcal L_{\rm fact}+0.25\mathcal L_{\rm bind}+0.25\mathcal L_{\rm edit}.
\label{eq:decodertraining}
\end{equation}
Edits add, remove or move components and carry updated factual targets. Duplicate candidate-query perturbations are also used during training. The frontend remains frozen; optimization and individual loss weights are specified in the supplementary material.

Background and padded slots are removed to obtain the object set. The center $\hat{\mathbf u}_i$ routes an identity to a witness; final measurement values are supplied by the association below.

\subsection{Partial one-to-one evidence association}
Let $E(x)=\{(c_j^E,\mathbf u_j^E,g_j^E)\}_{j=1}^{m}$ be the witness set, with $g_j^E$ containing center, box and axis. A decoded object and witness are eligible when their classes agree and their normalized center distance $d_{ij}=\|\hat{\mathbf u}_i-\mathbf u_j^E\|_2$ is at most $r$. Algorithm~\ref{alg:binding} constructs an $n\times(m+n)$ cost matrix with $n$ dummy columns and computes a minimum-cost injective assignment.

A matched object inherits the witness geometry with $a_i=1$; a dummy match sets $g_i=\varnothing$ and $a_i=0$. Both retain the decoded class and confidence, so unmatched identities remain counted. The indexed fact list preserves multiplicities, and the trace records decoded/witness indices and matching distance.

Normal development data select a single scheme from same-class or any-class association at radii 0.04 or 0.08. The selected same-class radius 0.08 is used across all categories and for both decoder variants.

\subsection{Support-certified geometric readout}
Initial witnesses use the frontend geometry, with the previously selected guarded native-axis reader for screws. This reader assigns box-supported screw-mask pixels and uses native-coordinate PCA; its support, centroid and elongation checks are specified in the supplementary material. Association fixes which initial witness belongs to each decoded identity. The subsequent reader refines the axis component of these bound objects, preserving their decoded identities, centers and boxes.

For a mask-backed length class $c$, let $I_c$ be its bound objects and $M_c$ the frozen foreground mask. Each mask pixel is assigned to the nearest bound center in native image coordinates, $\mathbf z_i=(Wu_{ix},Hu_{iy})$. Let $S_i$ be the pixels assigned to $i$, $\boldsymbol\mu_i$ their native-coordinate mean and $\mathbf v_i$ their principal direction. A candidate axis $\tilde g_i$ uses projection quantiles:
\begin{equation}
 \tilde{\mathbf a}_i^{\pm}=\boldsymbol\mu_i+
 Q_{.01/.99}\{(\mathbf p-\boldsymbol\mu_i)^\top\mathbf v_i:\mathbf p\in S_i\}\mathbf v_i.
 \label{eq:maskaxis}
\end{equation}
At least 16 pixels and finite, distinct endpoints are required. The candidate and initial axis are evaluated against the same $S_i$ using
\begin{equation}
 Q(g;S_i)=\sqrt{P_{\rm line}(g,S_i)P_{\rm extent}(g,S_i)}\,P_{\rm conn}(S_i),
 \label{eq:supportquality}
\end{equation}
where $P_{\rm line}$ is the supported fraction of sampled axis points, $P_{\rm extent}$ is the assigned-pixel fraction covered by the projected axis interval, and $P_{\rm conn}$ is the largest connected-region fraction. A candidate requires $Q(\tilde g_i;S_i)\ge q_0$ and a quality gain of at least $\Delta_q=0.05$. The global floor $q_0\simeq0.99087$ is the tenth percentile of readable candidate quality on normal development images.

High quality alone can admit an incomplete fragment. Let $K_c$ contain the substantial 8-connected components of the full class mask, and $k(i)$ be the component with greatest overlap with $S_i$. The class certificate is
\begin{equation}
\begin{aligned}
G_c={}&\ind[|K_c|=|I_c|]\,
\ind[k:I_c\!\to\!K_c\text{ is bijective}]\\
&\cdot\prod_{i\in I_c}\ind\!\left[\frac{|S_i\cap k(i)|}{|k(i)|}\ge0.98\right]
\ind[\mathbf u_i\in\mathcal D_2(k(i))],
\end{aligned}
\label{eq:completeownership}
\end{equation}
where $\mathcal D_2$ dilates a component by two mask pixels. Components smaller than $\max(16,\lceil0.02|M_c|\rceil)$ are excluded. The certificate connects the \emph{observed bound instances} to distinct class-mask components. These structural checks must pass for every instance before any axis of its class can be updated.

An additional guard $B_i$ rejects boundary-truncated components and unresolved shortening behind another object's box. The acceptance decision is
\begin{equation}
\begin{aligned}
\alpha_i={}&G_c B_i\,\ind[\tilde g_i\text{ readable}]\,
\ind[Q(\tilde g_i;S_i)\ge q_0]\\
&\cdot\ind[Q(\tilde g_i;S_i)-Q(g_i^0;S_i)\ge\Delta_q],\\
g_i={}&\begin{cases}\tilde g_i,&\alpha_i=1,\\g_i^0,&\alpha_i=0.\end{cases}
\end{aligned}
\label{eq:certifiedreadout}
\end{equation}
Here $g_i^0$ is the initial bound axis. Rejected updates retain the initial bound axis. The rule is shared across screw-bag, tape, rope, stick and stationery length readouts; other operands retain their original readers. Algorithm~\ref{alg:binding} gives the complete fact-construction procedure. The trace records both the initial witness link and any accepted mask readout or rejection reason.

\begin{table}[!t]
\centering\begin{spacing}{1.0}\small
\begin{paperalgorithm}{Binding and support-certified fact construction}\label{alg:binding}
\begin{tabularx}{\linewidth}{@{}r@{\hspace{6pt}}X@{}}
\textbf{Input} & Decoded objects $O$, initial witnesses $E$, class masks $M_c$; $r=0.08$, $q_0$, $\Delta_q=0.05$.\\[2pt]
1 & Set $C_{ij}=d_{ij}$ for same-class pairs with $d_{ij}\le r$; otherwise set $C_{ij}=10^6$. Add $n$ dummy columns of cost $r+10^{-7}$.\\
2 & $\pi\leftarrow\arg\min_{\pi\;\mathrm{injective}}\sum_i C_{i,\pi(i)}$ (Hungarian).\\
3 & Bind real matches to $g_i^0$; mark dummy matches unavailable. Retain every decoded identity.\\
4 & \textbf{for each} mask-backed length class $c$ with $I_c\ne\varnothing$ \textbf{do}\\
5 & \hspace{8pt}$S_i\leftarrow\{\mathbf p\in M_c:i=\arg\min_{h\in I_c}\|\mathbf p-\mathbf z_h\|_2\}$.\\
6 & \hspace{8pt}Where readable, compute $\tilde g_i$ by Eq.~\eqref{eq:maskaxis} and both qualities by Eq.~\eqref{eq:supportquality}.\\
7 & \hspace{8pt}Extract $K_c$, dominant components $k(i)$ and class certificate $G_c$ by Eq.~\eqref{eq:completeownership}.\\
8 & \hspace{8pt}Evaluate $B_i$; accept $\tilde g_i$ or retain $g_i^0$ by Eq.~\eqref{eq:certifiedreadout}. Record the support source and decision.\\
9 & \textbf{end for}\\
10 & $N_c\leftarrow\sum_i\ind[\hat c_i=c]$; assemble $f_i=(\hat c_i,p_i,g_i,a_i)$ and source trace $\mathcal T$.\\[2pt]
\textbf{Output} & Facts $\mathcal F$, counts $\{N_c\}$, measurement availability and evidence trace $\mathcal T$.\\
\end{tabularx}
\end{paperalgorithm}
\end{spacing}
\end{table}

\subsection{Constraint verification and normal calibration}
Discrete task requirements are encoded in the task's executable constraint configuration and evaluated on the predicted facts. Continuous intervals $[L_j,U_j]$ and scales $b_j$ are fitted from normal fitting/development observations. A measurement $m_j(x)$ yields
\begin{equation}
 r_j(x)=\max\{0,L_j-m_j(x),m_j(x)-U_j\}/b_j.
\end{equation}
Counts, lengths, locations and relations are reconstructed from the assembled objects, with category-specific postprocessing. Cached color, OCR, separator contrast and cable-centerline observations remain available to their task rules. Missing required geometry contributes an explicit unavailable-operand residual.

For normal fitting residual reference $R_{t,j}$, define
\begin{align}
 P_{t,j}(r)&=\frac{1+\sum_{v\in R_{t,j}}\mathbf1[v\ge r]}{|R_{t,j}|+1},\\
 z_{t,j}(x)&=\begin{aligned}[t]&-\log P_{t,j}(r_j(x))\\
 &+\frac{\max\{0,r_j(x)-\max R_{t,j}\}}{\max\{\operatorname{IQR}(R_{t,j}),1\}},\end{aligned}\\
 s_t(x)&=\max_jz_{t,j}(x)+0.01\sum_jz_{t,j}(x).
 \label{eq:score}
\end{align}
Independent normal calibration images $\mathcal C_t$ fix
\begin{equation}
\begin{aligned}
 \tau_t&=Q_{.95}^{\rm higher}\{s_t(x):x\in\mathcal C_t\},\\
 \hat y_t(x)&=\mathbf1[s_t(x)>\tau_t].
\end{aligned}
 \label{eq:decision}
\end{equation}
The strict comparison determines the decision at score ties. \mbox{Intervals}, residual references and calibration thresholds are frozen before inspection.

\subsection{Execution-grounded explanation}\label{sec:trace}
The executor records $\eta_j=(m_j,\rho_j,r_j,z_j,\mathcal P_j)$: observed operand, requirement, raw residual, normalized contribution and links to factual/visual sources. Counts cite decoded identities; measured predicates also cite the bound witness and accepted readout. Missing operands retain their availability state.

For an alarm, a compact certificate identifies a sufficient subset of these executed contributions. Let
\begin{equation}
 S(C)=\max_{j\in C}z_j+0.01\sum_{j\in C}z_j,\qquad S(\varnothing)=0.
 \label{eq:certscore}
\end{equation}
With $z_{(1)}\ge\cdots\ge z_{(J)}\ge0$, choose
\begin{equation}
 k^*=\min\{k:S(C_k)>\tau_t\},\quad C_k=\{(1),\ldots,(k)\}.
 \label{eq:tracecertificate}
\end{equation}
The descending prefix maximizes both terms of Eq.~\eqref{eq:certscore} at each cardinality, making $C_{k^*}$ a minimum-size sufficient certificate. Equal-score ties prefer available predicates and then key order. Omitted contributions are zeroed in this score-space definition.

A frozen language model realizes the selected predicate outcomes as source-attributed claims. For each selected record, its input contains the predicate name, operator, qualitative comparison and evidence ID. Names of unavailable operands are retained. Numeric operands, intervals and score comparisons remain in the executor's evidence records. We use the existing Qwen3-VL-8B-Instruct checkpoint in text-only mode \cite{qwen3}, with greedy decoding and frozen weights.

The model returns the recorded decision and claims paired with evidence IDs. A contract checks decision consistency, exact coverage of the selected and unavailable evidence, permitted IDs and explicit missingness; generated claims contain no numeric values. The executor attaches citations and renders the original numeric evidence beside the text. Outputs failing the contract retain a structured-trace report. The full trace preserves other deviations and unavailable operands, including in non-alarm reports. This connects fluent language to the same executed evidence used by the detector.

\section{Experiments}
\subsection{Datasets and information sources}
% Declared early so full-width benchmark tables stay beside experimental text.
\begin{widetable}
\centering\fontsize{8.5}{10}\selectfont\setlength{\tabcolsep}{2.3pt}
\renewcommand{\arraystretch}{1.04}
\caption{LOCO logical detection results (\%). Paired columns report AUROC and fixed-threshold F1. Mean averages five tasks. Supervision and implementation settings are specified in Section~\ref{sec:comparisons} and the supplement. Best/second-best distinct values are bold/underlined, with shared marks for ties. Blue columns denote SGDB.}
\label{tab:loco}
\begin{tabular}{@{}l*{6}{rr}>{\columncolor{OurBlue}}r>{\columncolor{OurBlue}}r@{}}
\toprule
 & \multicolumn{2}{c}{\shortstack{SALAD\\\cite{salad}}} & \multicolumn{2}{c}{\shortstack{LogiCo\\\cite{logico}}} & \multicolumn{2}{c}{\shortstack{ComAD\\\cite{comad}}} & \multicolumn{2}{c}{\shortstack{PUAD\\\cite{puad}}} & \multicolumn{2}{c}{\shortstack{Qwen3-VL-8B\\\cite{qwen3}}} & \multicolumn{2}{c}{\shortstack{LogicAD-FE-fast\\\cite{logicad}}} & \multicolumn{2}{c}{\shortstack{\textbf{SGDB}\\(Ours)}} \\
\cmidrule(lr){2-3} \cmidrule(lr){4-5} \cmidrule(lr){6-7} \cmidrule(lr){8-9} \cmidrule(lr){10-11} \cmidrule(lr){12-13} \cmidrule(lr){14-15}
Category & AUROC & F1 & AUROC & F1 & AUROC & F1 & AUROC & F1 & AUROC & F1 & AUROC & F1 & AUROC & F1 \\\midrule
Breakfast box & 92.27 & 85.00 & \textbf{98.89} & \underline{96.30} & 94.19 & 86.79 & 93.07 & 82.43 & 91.21 & 71.64 & 52.60 & 17.65 & \underline{98.78} & \textbf{98.78} \\
Juice bottle & \underline{99.79} & \textbf{98.61} & 99.22 & 95.62 & 87.96 & 79.68 & \textbf{99.83} & \textbf{98.61} & 83.31 & 61.46 & 53.67 & 1.40 & 96.38 & \underline{96.00} \\
Pushpins & \textbf{100.00} & 75.21 & 98.37 & \underline{93.55} & 87.11 & 78.53 & 86.69 & 69.44 & 62.84 & 14.29 & 53.02 & 8.00 & \underline{98.87} & \textbf{96.22} \\
Screw bag & \underline{93.95} & \underline{84.65} & 84.18 & 69.41 & 79.22 & 24.84 & 70.20 & 28.92 & 55.25 & 49.34 & 49.82 & 2.84 & \textbf{98.92} & \textbf{93.54} \\
Splicing connectors & \underline{96.08} & \textbf{88.78} & 91.80 & 84.31 & 85.61 & 82.90 & 94.30 & 88.00 & 78.56 & 56.95 & 65.70 & 8.85 & \textbf{97.97} & \underline{88.08} \\
\midrule
\textbf{Mean} & \underline{96.42} & 86.45 & 94.49 & \underline{87.84} & 86.82 & 70.55 & 88.82 & 73.48 & 74.23 & 50.74 & 54.96 & 7.75 & \textbf{98.19} & \textbf{94.52} \\
\bottomrule
\end{tabular}

\end{widetable}

\begin{widetable}
\centering\fontsize{8.5}{10}\selectfont\setlength{\tabcolsep}{2.3pt}
\renewcommand{\arraystretch}{1.04}
\caption{VID detection results by scenario (\%). Paired columns report AUROC and fixed-threshold F1. Rows average five conditions; Mean averages all 50 tasks. Supervision and calibration settings are specified in Section~\ref{sec:comparisons} and the supplement. Best/second-best distinct values are bold/underlined; blue columns denote SGDB.}
\label{tab:vidscenarios}
\begin{tabular}{@{}l*{6}{rr}>{\columncolor{OurBlue}}r>{\columncolor{OurBlue}}r@{}}
\toprule
 & \multicolumn{2}{c}{\shortstack{SALAD\\\cite{salad}}} & \multicolumn{2}{c}{\shortstack{LogiCo\\\cite{logico}}} & \multicolumn{2}{c}{\shortstack{ComAD\\\cite{comad}}} & \multicolumn{2}{c}{\shortstack{PUAD\\\cite{puad}}} & \multicolumn{2}{c}{\shortstack{Qwen3-VL-8B\\\cite{qwen3}}} & \multicolumn{2}{c}{\shortstack{LogicAD-FE-fast\\\cite{logicad}}} & \multicolumn{2}{c}{\shortstack{\textbf{SGDB}\\(Ours)}} \\
\cmidrule(lr){2-3} \cmidrule(lr){4-5} \cmidrule(lr){6-7} \cmidrule(lr){8-9} \cmidrule(lr){10-11} \cmidrule(lr){12-13} \cmidrule(lr){14-15}
Scenario & AUROC & F1 & AUROC & F1 & AUROC & F1 & AUROC & F1 & AUROC & F1 & AUROC & F1 & AUROC & F1 \\\midrule
Balls & 23.76 & 58.06 & 58.88 & 61.45 & 65.85 & 53.57 & 17.27 & 54.04 & \textbf{97.87} & \textbf{89.97} & 51.27 & 12.91 & \underline{86.43} & \underline{87.56} \\
Blocks & 90.92 & 84.62 & \textbf{94.37} & \textbf{90.98} & 77.93 & 71.64 & 79.70 & 74.55 & \underline{92.93} & \underline{90.37} & 48.78 & 14.69 & 89.15 & 86.98 \\
Cookies & 62.97 & 80.26 & 52.37 & 80.67 & \textbf{98.05} & \textbf{91.49} & 18.93 & 62.76 & 84.29 & \underline{88.03} & 54.24 & 10.32 & \underline{86.62} & 85.21 \\
Dishes & 52.36 & 37.59 & 66.61 & 73.74 & 73.95 & 57.72 & 61.84 & 43.60 & \textbf{99.64} & \underline{95.39} & 80.55 & 77.45 & \underline{92.90} & \textbf{96.63} \\
Fruits & \underline{99.81} & 94.60 & 99.17 & \underline{96.41} & \textbf{99.94} & \textbf{98.06} & 98.56 & 90.08 & 95.31 & 93.42 & 90.55 & 73.12 & 93.80 & 93.31 \\
Ropes & 43.00 & 3.67 & 31.93 & 5.26 & 80.20 & 29.63 & 41.65 & 4.19 & \underline{81.46} & \underline{77.46} & 58.45 & 30.33 & \textbf{88.50} & \textbf{86.33} \\
Stationery & 51.07 & \underline{76.85} & 62.29 & 76.16 & 65.61 & 58.43 & 42.39 & 63.48 & \underline{71.91} & 24.55 & 55.90 & 16.98 & \textbf{100.00} & \textbf{100.00} \\
Sticks & 82.90 & 23.83 & 45.63 & 41.94 & \underline{97.55} & \underline{88.28} & 82.16 & 22.00 & 85.27 & 78.46 & 69.12 & 16.82 & \textbf{99.92} & \textbf{98.77} \\
Tapes & \textbf{100.00} & 92.08 & 98.65 & 90.34 & \textbf{100.00} & \underline{94.35} & \textbf{100.00} & 88.69 & 92.94 & 71.35 & 68.97 & 29.00 & \underline{98.73} & \textbf{94.99} \\
Tools & 90.16 & 37.87 & \underline{93.88} & \underline{84.44} & 76.92 & 58.99 & 78.99 & 15.68 & 82.81 & 62.05 & 61.59 & 28.74 & \textbf{99.50} & \textbf{97.84} \\
\midrule
\textbf{Mean} & 69.70 & 58.94 & 70.38 & 70.14 & 83.60 & 70.22 & 62.15 & 51.90 & \underline{88.44} & \underline{77.11} & 63.94 & 31.04 & \textbf{93.56} & \textbf{92.76} \\
\bottomrule
\end{tabular}

\end{widetable}

We evaluate five MVTec LOCO AD categories using normal images and logical anomalies, for 1,136 test images; structural anomalies are excluded. VID-AD contributes 7,895 images across ten scenarios and five capture conditions, giving all 50 deployment tasks \cite{loco,vid}. Parent-image roles and known cross-condition source groups are preserved.

Normal images have separate fitting, development and calibration roles. Fitting observations train the readers and provide residual references; development observations select checkpoints/readout settings and pad continuous intervals; calibration scores fix alarm thresholds. LOCO contains 1,183 fitting, 147 development and 305 calibration images. VID uses 1,000, 500 and 500 images in these roles, with ten calibration images per deployment task.

Component annotations supply class, center, box, axis and weak-mask targets for visual learning. They cover all LOCO fitting/development parents and 194 fitting/147 development VID parents. Factual edits of fitting parents carry updated targets and can produce logically abnormal configurations. Manually specified discrete task requirements configure the executor; continuous measurement intervals are fitted from normal observations. Calibration images are independent of network and association selection. Method development included inspection of the public test sets; the reported results use the final frozen configurations. Supplementary Table~S1 gives the category-level budgets.

\subsection{Implementation and comparisons}\label{sec:comparisons}
The visual frontend is trained for 40 epochs per category at $448\times320$ resolution, using AdamW \cite{adamw}. Each semantic decoder adds 768 updates and 1,536 image exposures with the frontend frozen. At most 64 candidate queries and 64 grid queries pass through a width-64, four-head, two-layer decoder. Association is selected on 294 annotated normal development parents using the fixed point-F1/axis-error objective. The selected same-class radius is 0.08. The support-readout quality floor is fixed from 609 readable normal-development axes. The component and pixel tolerances are shared across categories; normal TRAIN/DEV/CAL data receive the same readout as test images. Architecture, optimization and readout details are provided in the supplementary material.

The external comparisons report the completed local deployments of SALAD, LogiCo, ComAD, PUAD, Qwen3-VL-8B and LogicAD-FE-fast. LOCO ComAD uses DINO-S8 clustering and CRF without component labels; its VID deployment uses a shared component-map frontend. Local PUAD omits the external ImageNet penalty-image term, and LogicAD-FE-fast uses a simplified program/frontend configuration. VID SALAD and PUAD use category-level calibration; SGDB uses task-level calibration. The supplementary comparison-condition table and the archive condition matrix specify their supervision, implementation and calibration settings. AUROC and fixed-threshold F1 are percentages, macro-averaged over five LOCO tasks or all 50 VID tasks.

Internal controls share normal-source partitions, component-supervision budgets, the visual frontend and task requirements. Decoder variants use the same update budget. Binding/readout controls reuse frozen semantic outputs and visual witnesses. Each configuration fits its references and threshold from its own normal facts using the same scoring and calibration recipe; the readout study also fixes the initial references and thresholds to isolate measurement changes.

\subsection{Detection performance}
Tables~\ref{tab:loco} and \ref{tab:vidscenarios} report complete category/scenario comparisons. SGDB achieves 98.19 AUROC and 94.52 F1 on LOCO logical, and 93.56 AUROC and 92.76 F1 on VID. Relative to the FMP plus guarded-geometry reference, the changes are +0.11/+0.98 percentage points on LOCO AUROC/F1 and +2.22/+3.30 on VID. The VID result reflects the complete set of capture conditions.

Figures~\ref{fig:visualcomparison} and \ref{fig:vidvisualcomparison} compare 20 LOCO and 30 VID sources. Source order is fixed lexicographically across categories and conditions. Each prediction strip shows the decision and its score/threshold comparison.

SGDB's boxes and axes come from its support-checked associated objects; the semantic colors depict the frozen frontend witnesses. Baseline representations retain their original meanings: SALAD shows saved local appearance/composition maps and ComAD displays component regions.

The disaggregated results show complementary strengths. SGDB attains the highest displayed LOCO AUROC on screw bag and splicing connectors, while category leaders differ across the benchmark. On VID, support-certified readout raises Tapes from 89.43/85.27 to 98.73/94.99 AUROC/F1, alongside strong results for Stationery, Sticks and Ropes. The fixed-threshold normal FPR is 1.43\% on LOCO and 11.36\% on VID. Complete task-level results and condition summaries appear in the supplementary material.

\gdef\firstcasefigure{\begin{minipage}[b]{\textwidth}\centering\captionsetup{type=figure}
\includegraphics[width=\linewidth]{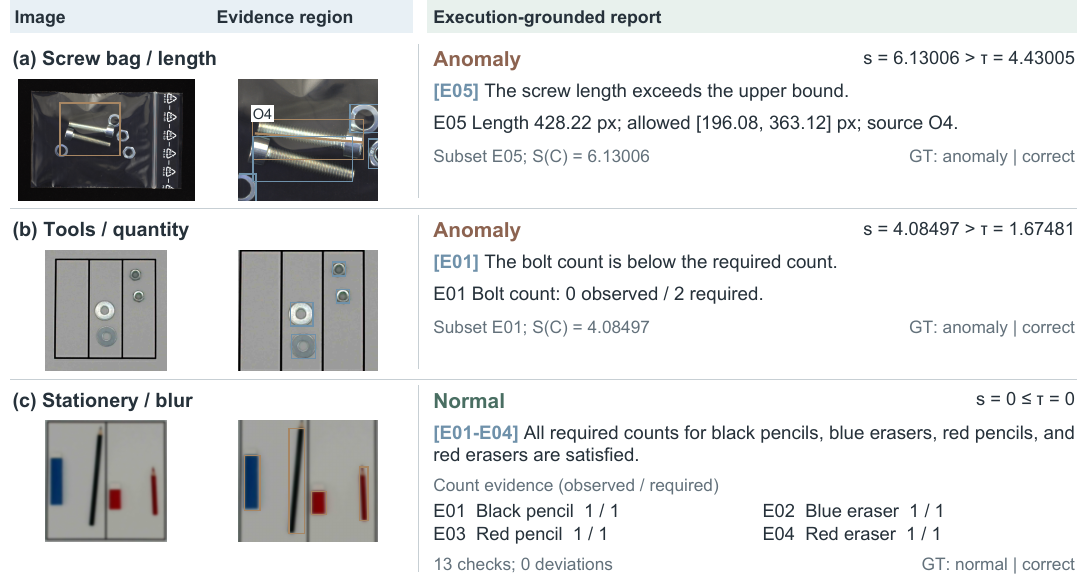}
\caption{Execution-grounded reports: (a) screw-length alarm; (b) missing-bolt alarm; (c) normal Stationery under blur. Rows separate visual evidence, decisions, frozen-LLM claims and executor-owned measurements. An alarm requires $s>\tau$; $S(C)$ is the selected subset score. Evidence IDs are local to each case. GT labels are used only for retrospective review.}
\label{fig:ebfdcases}
\end{minipage}}

% Frozen neural outputs and completed baseline results are reused.
\subsection{Mechanism controls under shared supervision}
Table~\ref{tab:ablation}(a) separates semantic decoding, initial witness readout and final support certification under the shared setup above. Direct evidence-edit DFD and its initial bound readout use the same decoder predictions and task requirements. Replacing direct geometry with bound witnesses raises LOCO AUROC/F1 from 95.42/92.41 to 98.19/94.52; screw-bag AUROC rises from 85.04 to 98.92. The certificate then improves VID from 92.63/91.79 to 93.56/92.76 while preserving LOCO performance. The plain-set hybrid provides a decoder control without the certificate.

\begin{table}[!htb]
\centering\footnotesize\setlength{\tabcolsep}{3pt}\begin{spacing}{1.0}
\caption{Mechanism controls with shared component supervision and task requirements (\%). (a) Decoder/readout configurations; the final row adds the support certificate. (b) Binding policies with the same complete-support readout. Each row uses its own normal references and CAL threshold under the shared recipe. Best/second-best distinct values are bold/underlined within each block and column; ties share marks.}
\label{tab:ablation}
\begin{tabularx}{\columnwidth}{@{}Xrrrr@{}}\toprule
&\multicolumn{2}{c}{LOCO logical}&\multicolumn{2}{c}{VID all 50}\\
Configuration & AUROC & F1 & AUROC & F1\\\midrule
\multicolumn{5}{@{}l}{\textit{(a) Semantic decoder and geometric readout}}\\[2pt]
\RowsEBFDAblation
\midrule
\multicolumn{5}{@{}l}{\textit{(b) Binding-policy controls}}\\[2pt]
\RowsBindingPolicies
\bottomrule\end{tabularx}\end{spacing}
\end{table}

Table~\ref{tab:ablation}(b) changes binding or unbound-object handling while retaining the certified readout. Nearest assignment allows witness reuse; the other controls remove class compatibility or distance gating, drop unbound identities, or tighten the radius. Removing class compatibility gives 92.60/91.38 on VID with its own references, versus 93.55/92.76 with shared SGDB references. Nearest assignment and removal of the distance gate leave detection metrics unchanged. Radius 0.04 gives 97.98/94.52 on LOCO and 93.97/93.11 on VID; radius 0.08 remains the normal-development selection.

Controlled removal of each matched witness on the 294 annotated normal development parents gives 3,136 binding-stage trials. The semantic identities and remaining witnesses stay fixed. Partial binding retains decoded counts and marks orphan geometry unavailable in every trial. Nearest assignment reuses another witness in 720 trials (22.96\%); dropping unbound objects changes every trial's count. This probe measures identity preservation, missing-evidence representation and witness reuse at the binding interface. Complete results appear in the supplementary material.

\subsection{Support-certified readout}
Table~\ref{tab:readout} changes only the axis-update rule, holding frozen semantic outputs, initial witnesses, masks and task requirements fixed. Uniform mask readout improves VID but lowers LOCO AUROC to 96.36. Quality selection alone can admit an incomplete, internally clean fragment. Complete component ownership retains all 90 tape corrections while preventing the screw degradations introduced by quality-only readout. The complete-support deployment corrects 90 prior errors and introduces one type-related tape miss, for 89 fewer errors across all 9,031 test images.

\begin{table}[!htb]
\centering\footnotesize\setlength{\tabcolsep}{3pt}\begin{spacing}{1.0}
\caption{Geometric-readout controls with fixed semantic outputs, witnesses, masks and task requirements (\%). Normal references and thresholds follow the same fitting/calibration recipe. SGDB combines quality selection with complete instance support. Best/second-best distinct values are bold/underlined; ties share marks.}
\label{tab:readout}
\begin{tabularx}{\columnwidth}{@{}Xrrrr@{}}\toprule
&\multicolumn{2}{c}{LOCO logical}&\multicolumn{2}{c}{VID all 50}\\
Readout & AUROC & F1 & AUROC & F1\\\midrule
\RowsReadoutAblation
\bottomrule\end{tabularx}\end{spacing}
\end{table}

Fixing the initial intervals, references and thresholds gives the same 90 corrections and one new tape miss, attributing these transitions to measurement selection. The completeness check filters 20 of 23 quality-eligible screw axes and preserves all 259 screw decisions, scores and thresholds. The tape corrections cover 20 cross-condition filename-index groups. Support coverage and per-source transitions are provided in the supplementary material.

Paired resampling of 10,000 category--label-stratified groups gives V2-minus-initial VID gains of 0.93 AUROC points (95\% interval 0.56--1.32) and 0.97 F1 points (0.57--1.43). The five capture conditions travel together within filename-index groups. V2's LOCO AUROC advantages over uniform and quality-only readout are 1.83 points (1.18--2.51) and 1.57 points (0.96--2.22). Uniform readout retains a small VID F1 advantage: the V2-minus-uniform difference is $-0.10$ points ($-0.20$ to $-0.02$). The paired intervals and source-index transition ledgers in the supplement locate the benefit in preserving reliable geometry across categories. These conditional intervals use the existing, previously inspected test cohorts.

\subsection{Execution-grounded explanation}\label{sec:explanationexp}
We evaluate evidence selection on the same frozen V2 deployment, retaining its facts, references, thresholds and all 9,031 predictions. The 66,297 executed rule records provide operands and source links. Table~\ref{tab:explanation} compares four policies on all 5,725 alarms. Sufficiency is the fraction whose selected contributions still exceed the original threshold under Eq.~\eqref{eq:certscore}; mean size counts selected rule records.

\begin{table}[!htb]
\centering\footnotesize\setlength{\tabcolsep}{3pt}\begin{spacing}{1.0}
\caption{Explanation selection on 5,725 frozen alarms. Mean: selected rules per alarm; Suff.: alarm sufficiency (\%); Fail: insufficient selections. All policies use the same trace and source links. Bold/underline mark best/second-best sufficiency and failure values; ties share marks.}
\label{tab:explanation}
\begin{tabularx}{\columnwidth}{@{}Xrrr@{}}\toprule
Selection & Mean & Suff.$\uparrow$ & Fail$\downarrow$\\\midrule
\RowsExplanation
\bottomrule\end{tabularx}\end{spacing}
\end{table}

The compact certificate retains sufficiency for every alarm with 1.006 rules on average, reducing the selected count by 49.56\% relative to all positive-score rules. A single maximum-score rule is insufficient for 17 tape alarms: two strongest contributions give 4.125413, exactly the threshold, whereas three give 4.165858 and trigger the strict comparison. These are jointly sufficient score contributions, including the unavailable-measurement term.

Reports preserve missing evidence in all 1,254 affected records, including 94 non-alarms. Raw predicate deviations occur in 191 non-alarm traces, so their reports distinguish deviation from an alarm. Source checks reproduce all scores and decisions; an independent calculation reconstructs 3,928 cited geometric operands with maximum absolute error $5.7\times10^{-14}$.

The frozen language model expresses selected predicate outcomes as source-attributed claims, paired with executor-rendered numeric evidence in Figs.~\ref{fig:ebfdcases} and \ref{fig:ebfdlimits}. Per-category results, certificate checks and language-output consistency results are provided in the supplementary material.

\subsection{Decision, explanation and evidence}\label{sec:qualitative}
\begin{figure*}[p]
\centering
\includegraphics[width=.95\linewidth]{figures/ebfd_cases_main.pdf}
\caption{Execution-grounded reports: (a) screw-length alarm; (b) missing-bolt alarm; (c) normal Stationery under blur. Rows separate visual evidence, decisions, frozen-LLM claims and executor-owned measurements. An alarm requires $s>\tau$; $S(C)$ is the selected subset score. Evidence IDs are local to each case. GT labels are used only for retrospective review.}
\label{fig:ebfdcases}
\par\vspace{3pt}
\includegraphics[width=.95\linewidth]{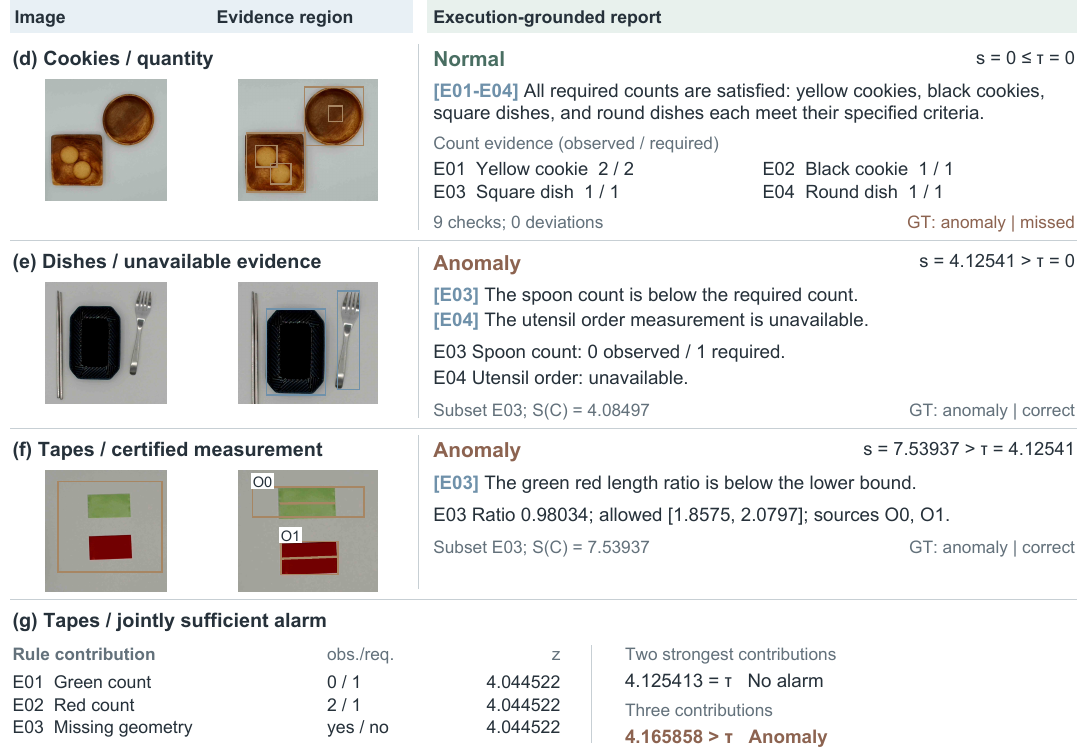}
\caption{Reports for (d) a Cookies miss, (e) missing spoon with unavailable order, and (f) support-certified tape measurement. Panel (g) separates the rule contributions from their sufficient subsets: two contributions equal the threshold without triggering an alarm, whereas three exceed it. Columns and symbols follow Fig.~\ref{fig:ebfdcases}; numeric evidence comes from the execution traces.}
\label{fig:ebfdlimits}
\end{figure*}

Figures~\ref{fig:ebfdcases} and \ref{fig:ebfdlimits} render six reports with language-model claims and executor-owned numeric evidence. Each report pairs generated claims with the full score comparison, selected rule contributions and source-linked operands. The screw report selects the short-length predicate: 428.22 pixels exceeds the 363.12-pixel bound, giving a sufficient certificate score of 6.13006. The Tools report selects zero decoded bolts against a requirement of two. The blurred Stationery report cites its evaluated counts and retains the complete zero-residual trace.

Cookies receives a zero-residual normal report despite its anomalous image label, exposing the available semantic facts behind the miss. Dishes cites the missing-spoon count while preserving unavailable utensil order. In Tapes, the accepted support readout changes the green/red length ratio from 2.053 to 0.980, below its interval; the report links that predicate to the two measured axes and their support records. Panel (g) of Fig.~\ref{fig:ebfdlimits} additionally shows why the joint tape alarm requires three score contributions rather than a single violated predicate.

% Queue benchmark plates after the small ablation table to keep it beside its text.
\begin{widefigure}
\includegraphics[width=\linewidth]{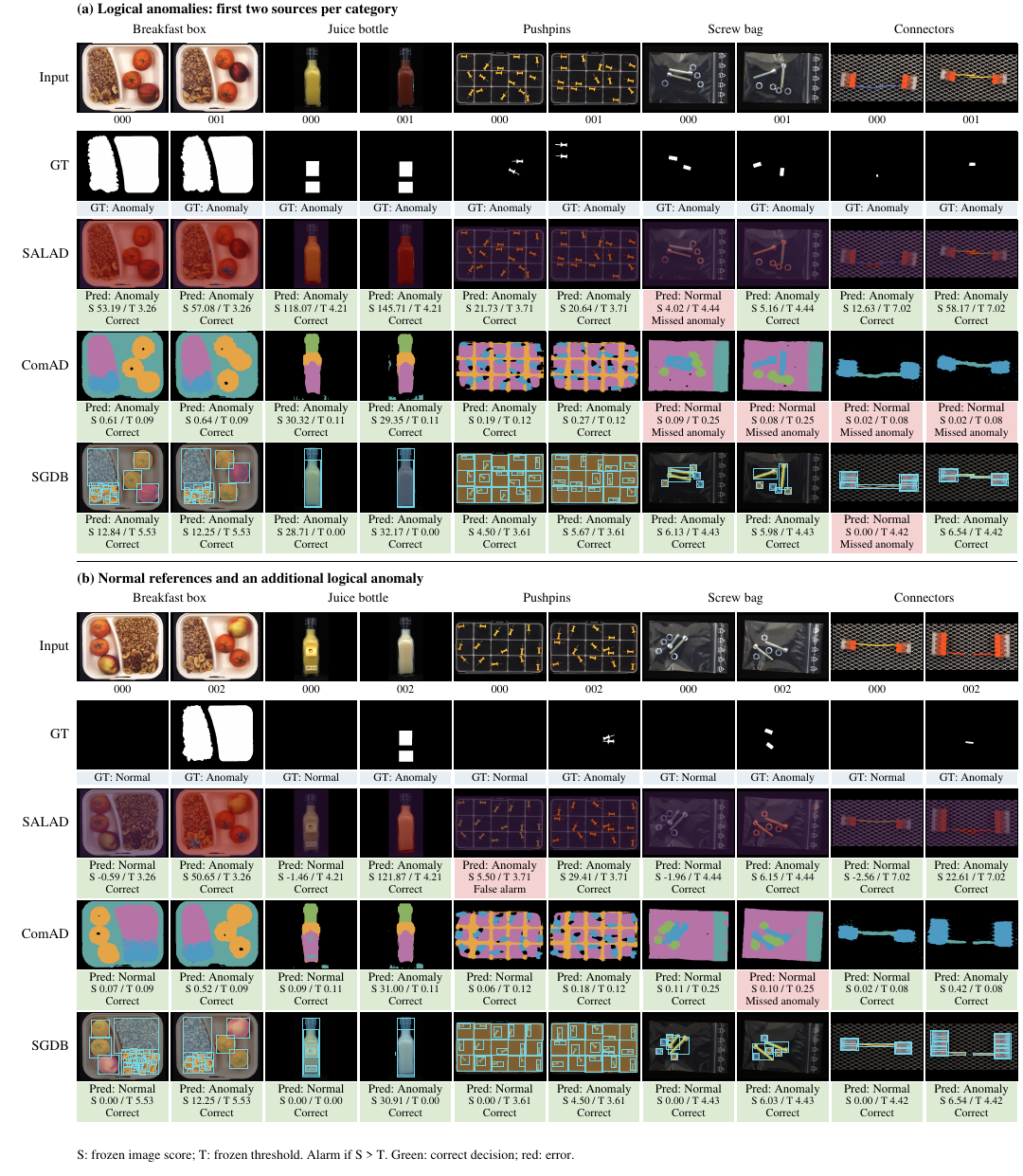}
\caption{LOCO comparison on 20 fixed sources: (a) logical anomalies; (b) normal references and an additional logical anomaly. GT is the provided logical-anomaly mask union. Prediction strips use full-precision scores; green/red denotes a correct/incorrect image-level decision. Spatial representations differ across methods.}
\label{fig:visualcomparison}
\end{widefigure}

\begin{widefigure}
\includegraphics[width=\linewidth,height=.88\textheight,keepaspectratio]{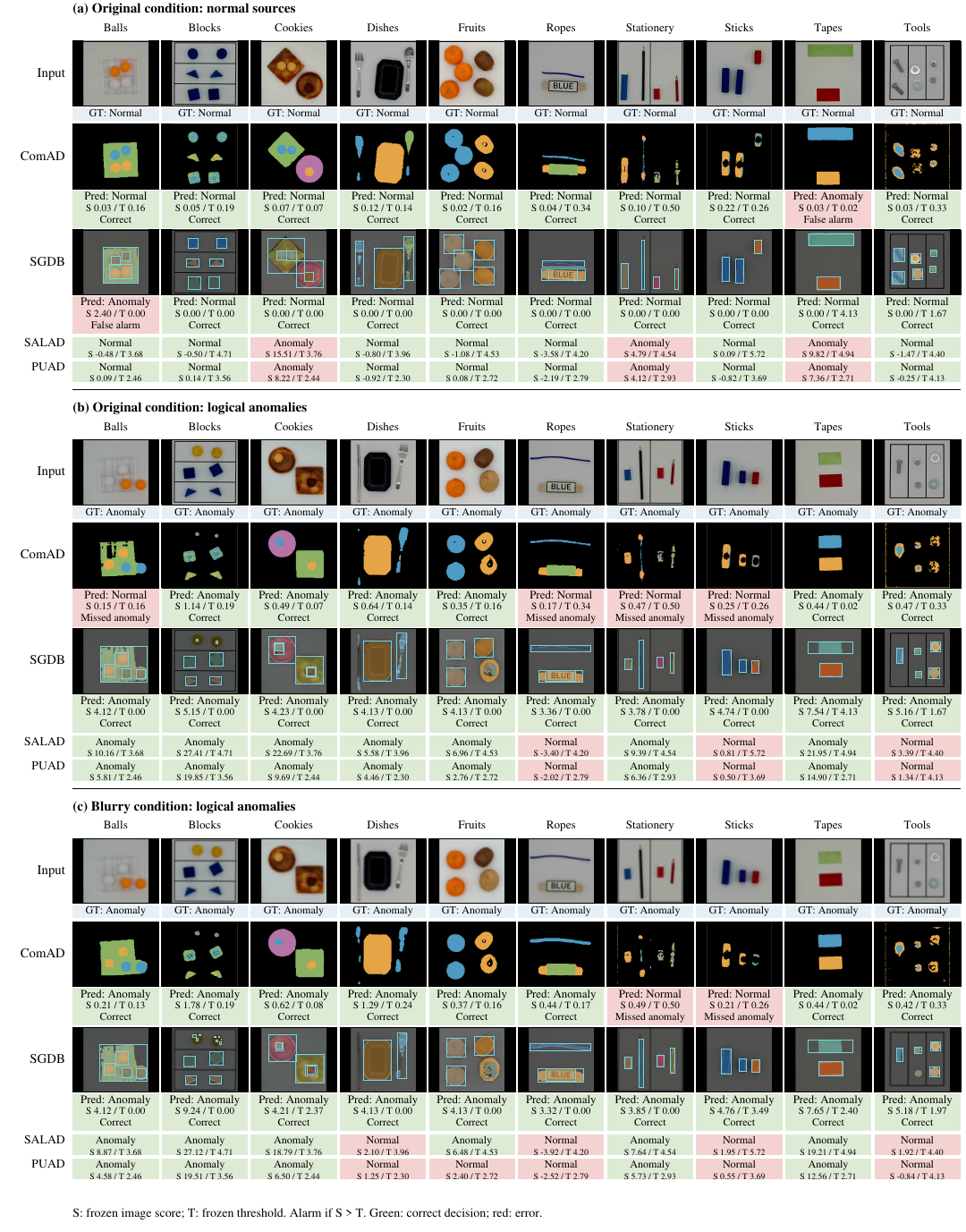}
\caption{VID comparison on 30 fixed sources across ten scenarios: (a) original normal, (b) original anomalous, and (c) blurry anomalous images. GT is image-level. SGDB shows associated objects; ComAD shows component regions. SALAD and PUAD provide scored decisions from their cached outputs.}
\label{fig:vidvisualcomparison}
\end{widefigure}

% Two separately numbered case figures share one float page.

% Flush the three full-width results plates before the end matter.
\clearpage
\makeatletter
\global\@colht\textheight
\global\@colroom\textheight
\global\vsize\textheight
\global\@firstcolumntrue
\makeatother
\twocolumn
\section{Discussion}
SGDB connects semantic identities, geometric measurements and executable requirements through a common evidence representation. Bound witness geometry improves length verification over direct prediction; complete-support selection retains tape corrections and protects screw measurements. Partial binding preserves decoded counts, unavailable geometry and witness ownership. Execution-grounded explanation carries these source links into a compact account of the score comparison, including alarms that require several contributions.

\section{Conclusion}
SGDB decouples semantic recognition from geometric measurement and reconnects them through partial evidence binding and support-certified readout. The resulting fact and source trace supports both executable verification and compact sufficient alarm certificates. Reports connect selected predicates to object identities, measurements and evidence availability. SGDB achieves 98.19/94.52 AUROC/F1 on LOCO logical and 93.56/92.76 on all 50 VID tasks. Shared-supervision and frozen-output controls examine the roles of geometric evidence, complete support and execution-grounded explanation.

Normal observations define measurement and score references, component annotations train the visual readers, and task requirements configure the executor. Extending this setup to unfamiliar scenes and reduced component supervision is a useful direction. The Cookies trace illustrates another role of the report: an anomalous image can yield zero-residual semantic facts, making the representation behind a missed alarm available for review.

% Keep the short end-matter headings compact without changing body typography.
\begingroup
\makeatletter
\renewcommand{\section}{\@startsection{section}{1}{\z@}{4pt plus 1pt minus 1pt}{2pt}{\normalfont\normalsize\bfseries}}
\makeatother
\fontsize{9.5}{10.2}\selectfont
\section*{CRediT authorship contribution statement}
\textbf{Chenrui Ji:} Conceptualization, Methodology, Software, Investigation, Writing -- original draft. \textbf{Jiafan Lu:} Software, Validation, Visualization, Writing -- review \& editing. \textbf{Zilu Luo:} Validation, Writing -- review \& editing. \textbf{Feng Wan:} Formal analysis, Validation, Writing -- review \& editing. \textbf{Qiuhui Chen:} Conceptualization, Methodology, Supervision, Project administration, Writing -- review \& editing. \textbf{Weimin Zhong:} Supervision, Resources, Funding acquisition, Writing -- review \& editing.
\section*{Data and code availability}
MVTec LOCO AD and VID-AD are available from their original providers \cite{loco,vid}. The research code, frozen configurations, per-image outputs and CPU reproduction scripts are publicly available at \url{https://github.com/Qybc/SGDB}, release \href{https://github.com/Qybc/SGDB/releases/tag/v1.0.0}{v1.0.0}. The release supports metric reconstruction, execution-trace replay and recorded language-output verification. Benchmark images and third-party pretrained models are obtained from their original providers.
\section*{Declaration of competing interests}
The authors declare no competing interests.
\section*{Funding}
This work was supported in part by the Fundamental and Interdisciplinary Disciplines Breakthrough Plan of the Ministry of Education of China (Grant No. JYB2025XDXM402) and the National Natural Science Foundation of China (Grant Nos. 62633009 and 62603270).
\section*{Declaration of generative AI and AI-assisted technologies in manuscript preparation}
Codex assisted with manuscript editing and LaTeX preparation.
\bibliographystyle{elsarticle-num}
\bibliography{references}
\endgroup
\end{document}

\maketitle
\section{Data roles and implementation}
\subsection{Information sources and normal-data budgets}
SGDB uses normal images, component targets and explicit task requirements for distinct parts of inspection. Normal fitting images supply the original visual training sources and residual references. Normal development images select reader checkpoints, association/readout settings and continuous-interval padding. Independent normal calibration images fix the alarm thresholds. Same-parent edits inherit their source role, and known cross-condition source groups are preserved.

Component targets provide class, center, box, axis and weak-mask supervision for the visual readers. Table~\ref{tab:roles} distinguishes the full normal pools from the parents carrying these targets. At inspection, objects and masks are predicted from the query image. Discrete task requirements are encoded as executable constraints on the resulting facts; continuous bounds and residual scales are estimated from normal observations. The task identifier selects the relevant readers, constraints and normal references.

\begin{table}[tbp]
\centering\small\begin{spacing}{1.08}
\caption{Normal-source budgets and component-supervised parents. Ann.: annotated with component facts and weak masks.}
\label{tab:roles}
\begin{tabular}{lrrrrrr}\toprule
Category & Fitting & Dev. & Cal. & Ann. fit. & Ann. dev. & Test\\\midrule
\RowsRoleCounts
\bottomrule\end{tabular}\end{spacing}
\end{table}

\subsection{Architecture and optimization}
The visual frontend uses a WideResNet50-2 encoder truncated after the third residual stage. Global-context modulation, multiscale upsampling/skip fusion and explicit normalized coordinates feed center, semantic-mask, box and endpoint heads. Training uses
\begin{equation}
\mathcal L_F=\mathcal L_{\rm center}+3\mathcal L_{\rm seg}
 +3\mathcal L_{\rm box}+3\mathcal L_{\rm axis}.
\end{equation}
The segmentation term combines weighted binary cross-entropy and Dice loss on valid weak-mask channels. Endpoint order is handled by the minimum loss over the two permutations. Category frontends receive 40 epochs with 240 sampled examples per epoch and batch size four. AdamW uses encoder/head learning rates $10^{-4}/10^{-3}$, weight decay $10^{-4}$ and a cosine schedule. The encoder is frozen for the first two epochs and batch-normalization statistics stay frozen. The frontend seed is 42.

The DFD has 64 candidate slots and 64 grid slots, width 64, four attention heads and two layers. Grid slots use a fixed $8\times8$ layout. Spatially pooled feature memory has a $40\times56$ grid. First-layer soft support masks add a log-sigmoid bias to second-layer attention. The main decoder objective is
\begin{equation}
\mathcal L_D=\mathcal L_{\rm cls}+2\mathcal L_{\rm ctr}
+2\mathcal L_{\rm box}+3\mathcal L_{\rm axis}+\mathcal L_{\rm mask}
+0.25\mathcal L_{\rm bind}+0.25\mathcal L_{\rm edit}.
\end{equation}
Support coupling aligns geometric predictions with mask support. Paired-edit consistency uses count changes and shared-object displacement, with duplicate source-query perturbations. The plain-set control omits mask feedback and the coupling/consistency mechanisms. Each head receives 768 optimizer steps and 1,536 image exposures. Fifteen categories are trained for each of the two variants.

Fitting data comprise normal-source parents and paired edits with updated factual targets. An edit can produce a logically abnormal configuration; supervision remains component/fact based. The decoder's geometric predictions support its training objective, while final hybrid measurements use associated witnesses.

\section{Association and geometric-readout selection}
\subsection{Normal-development association}
Four candidates combine same-class or any-class eligibility with radii 0.04 or 0.08. Selection uses 294 annotated original normal development parents across 15 categories. For each category, the objective is
\begin{equation}
0.8F_{1,\rm point}+0.2(1-E_{\rm axis}),
\end{equation}
where relative matched-axis error is clipped to $[0,1]$. The category objectives are macro-averaged. Unbound recognized objects count as false positives and unmatched reference objects as false negatives. Ties follow the preset candidate order. The same selected scheme is applied to both datasets and decoder variants.

\begin{table}[tbp]
\centering\small\begin{spacing}{1.05}
\caption{Complete normal-development association selection. Point F1 and clipped axis error are category macro-averages (\%). Unbound is a count.}
\label{tab:association}
\begin{tabular}{lrrrr}\toprule
Candidate & Point F1 & Axis error & Unbound & Objective\\\midrule
\RowsAssociationSelection
\bottomrule\end{tabular}\end{spacing}
\end{table}

The selected same-class radius 0.08 has 3,136 decoded development objects and no unbound objects. The main hybrid's test set has one unbound object in one breakfast-box image. The null policy retains that identity in the count and leaves its geometry-dependent operands unavailable.

\subsection{Initial screw-mask ownership and native axes}
For instance $i$, predicted box width and height $\mathbf w_i$ are lower-bounded by 0.02. Foreground pixels are eligible within the box expanded by 12.5\% on each side. Among eligible instances, a pixel $\mathbf p$ is assigned by minimizing
\begin{equation}
d_i(\mathbf p)=\|(\mathbf p-\mathbf u_i)\oslash\mathbf w_i\|_2^2.
\end{equation}
Unsupported pixels are excluded. Assigned pixels are mapped to native coordinates before PCA, and the selected reader uses untrimmed projection endpoints.

The substitution requires at least 16 assigned pixels, 75\% expanded-box support, normalized centroid displacement at most 0.04, principal-to-secondary eigenvalue ratio at least four, and finite endpoints. Otherwise the original FMP axis is retained. These fallback axes remain valid witnesses for subsequent identity association. A dummy semantic-to-witness match instead produces null geometry.

The readout candidates combine connected-component or box-supported assignment with zero or 1\% projection trimming. Selection uses 30 normal development screw-bag images with 60 screw instances. Admission requires at least a 10\% mean-error reduction, no regression in median, P95 or errors above 20\%, and at least one valid substitution. Candidate order resolves remaining ties after mean and P95 error.

\begin{table}[tbp]
\centering\small\begin{spacing}{1.05}
\caption{Native-axis error on 60 normal development screw instances. Relative errors are percentages; Large counts errors above 20\% and Updates counts substituted axes. Reference axes are stored normal component targets.}
\label{tab:geometry}
\begin{tabular}{>{\raggedright\arraybackslash}p{65mm}rrrrr}\toprule
Readout & Mean & Median & P95 & Large & Updates\\\midrule
\RowsGeometryDev
\bottomrule\end{tabular}\end{spacing}
\end{table}

\subsection{Post-binding support certification}
The initial witnesses above are bound to decoded identities before the final readout. For screw bags, tapes, ropes, sticks and stationery, same-class foreground pixels are reassigned to the nearest bound center using Euclidean distance in native coordinates, without clipping to boxes. At least 16 assigned pixels are required. PCA directions use native-coordinate covariance; endpoints use the 1st and 99th projection percentiles.

Quality is evaluated for the initial and candidate axes on the same assigned support. $P_{\rm line}$ samples 64 equally spaced axis points and measures membership in the assigned binary mask dilated once with a $3\times3$ kernel. $P_{\rm extent}$ measures how many assigned pixels project inside the axis interval padded by one native mask-pixel diagonal. $P_{\rm conn}$ is the largest 8-connected assigned-region area divided by all assigned pixels. The fixed quality floor is 0.9908659040240699, the tenth percentile over 609 readable normal-development candidates; the required gain is 0.05.

Substantial components of the full class mask have at least $\max(16,\lceil0.02|M_c|\rceil)$ pixels. Their count must equal the number of currently bound recognized instances, and dominant components must be distinct. Each instance must retain at least 98\% of its full dominant component and its frozen center must lie within a two-mask-pixel dilation of that component. Structural failure rejects every proposed update in that class. These checks use observed identities, without replacing them by normal expected counts.

The boundary guard rejects a dominant component touching the image edge or its adjacent one-pixel band. If a candidate shortens the initial axis by more than one native mask-pixel diagonal, 64 initial-axis samples are projected onto the candidate direction. Samples outside the padded candidate interval, together with its two endpoints, are tested against every other recognized object's box expanded by one mask pixel. Any intersection rejects the update. The rejection retains the initial axis, its witness link and availability; it does not delete an object or supply an expected normal length. Accepted updates retain the original identity-to-witness link and additionally record the mask source, assigned support, candidate quality and gate results.

The tolerances were fixed before the completeness experiment. The same reader is applied to normal fitting/development/calibration and test images, without category-specific switches or selection by proximity to a normal length interval. The development setting is summarized in Sec.~4.1 of the main paper.

\begin{table}[tbp]
\centering\small\begin{spacing}{1.05}
\caption{Certified readout coverage on the 3,429 test images with length readouts. Axes: bound target axes; Quality: candidates passing the quality gate; Accepted: candidates passing the complete certificate.}
\label{tab:supportcoverage}
\begin{tabular}{lrrr}\toprule Category & Axes & Quality & Accepted\\\midrule
Screw bag & 490 & 23 & 3\\
Tapes & 1,911 & 200 & 172\\
Ropes & 1,627 & 9 & 0\\
Sticks & 2,372 & 6 & 1\\
Stationery & 3,202 & 2 & 0\\
\bottomrule\end{tabular}\end{spacing}
\end{table}

The certificate updates 176 axes in 175 test images. With own normal references or the original bound deployment's references shared, all final decisions are identical between the two modes. The tape repair set is the same 90 images, corresponding to 20 cross-condition filename-index groups, with one additional type-related miss. All screw-bag predictions and normal configuration equal the initial bound deployment. The complete-support and whole-component-only variants yield exactly the same final facts and predictions, so the extra boundary/shortening guards have no measured independent benefit here.

\section{Frozen-output binding-policy ablations}
\subsection{Policies and evaluation modes}
All six policies reuse the evidence-edit decoder's frozen semantic outputs and the same initial FMP/guarded-screw witnesses, followed by the fixed complete-support readout. The main policy is the previously selected same-class, one-to-one assignment at radius 0.08. Independent nearest assignment uses the same eligibility test but allows several decoded objects to use one witness. The any-class control removes only class compatibility. The ungated control allows every same-class pair, with dummy cost $\sqrt{2}+10^{-7}$, so only witness scarcity requires an unbound object. The drop-unbound control first applies the main assignment and then discards objects assigned to a dummy. The radius control uses 0.04 with all other main-policy choices retained. The six binding policies and all readout tolerances are retained without a new search.

For each policy, original normal fitting and development parents rebuild the intervals and residual references; independent calibration parents set the 95th-percentile threshold using the same strict comparison. Auxiliary models, source roles, task rules and the score formula are unchanged. A second diagnostic applies the original SGDB references and thresholds to every policy's facts. This separates a changed fact interface from the additional effect of normal-reference fitting and calibration. No network forward, optimizer update or baseline execution is performed.

\begin{table}[tbp]
\centering\footnotesize\setlength{\tabcolsep}{2.8pt}\begin{spacing}{1.05}
\caption{Binding policies with their own normal references and CAL thresholds, and with SGDB's references/thresholds shared (\%). Each four-column group reports LOCO AUROC/F1, then VID AUROC/F1.}
\label{tab:bindingmodes}
\begin{tabularx}{\linewidth}{@{}Xrrrrrrrr@{}}\toprule
&\multicolumn{4}{c}{Own normal references}&\multicolumn{4}{c}{Shared SGDB references}\\
&\multicolumn{2}{c}{LOCO}&\multicolumn{2}{c}{VID}&\multicolumn{2}{c}{LOCO}&\multicolumn{2}{c}{VID}\\
Policy&AUC&F1&AUC&F1&AUC&F1&AUC&F1\\\midrule
\RowsBindingModes
\bottomrule\end{tabularx}\end{spacing}
\end{table}

The class-agnostic policy gives 92.60/91.38 VID AUROC/F1 with its own normal references and 93.55/92.76 with shared SGDB references. The tighter radius gives 93.97/93.11 on VID and 97.98/94.52 on LOCO. These controls retain the original normal-development choice of radius 0.08.

\begin{table}[tbp]
\centering\small\begin{spacing}{1.05}
\caption{Binding traces on all 9,031 test images and 51,158 decoded objects. Unbound counts decoded objects without a witness; Dropped counts identities removed from the assembled facts. Reuse counts assignments beyond one use per witness. Cross-class counts links with different decoded/witness labels.}
\label{tab:bindingtraces}
\begin{tabularx}{\linewidth}{@{}Xrrrr@{}}\toprule
Policy & Unbound & Dropped & Reuse & Cross-class\\\midrule
\RowsBindingTraces
\bottomrule\end{tabularx}\end{spacing}
\end{table}

Independent nearest assignment changes 3 test scores and 0 decisions. Dropping unbound identities changes 1 scores and 0 decisions. Removing the distance gate changes 0 scores. Binding-trace counts are unchanged by the downstream axis readout.

\subsection{Controlled missing-witness probe}
The binding-stage probe, before the downstream support readout, uses all 294 annotated original normal development parents. For every originally matched decoded object, its assigned witness is removed once, producing 3,136 trials. The image, semantic output, remaining witnesses and their measurements are unchanged. Every policy is then reapplied. These modified witness sets are used only to inspect binding behavior; they do not fit normal references or thresholds, select a configuration, or represent additional real anomaly images.

\begin{table}[tbp]
\centering\small\begin{spacing}{1.05}
\caption{Controlled witness removal over 3,136 trials (\%). Count kept: all decoded identities remain counted. Null orphan: the decoded object whose original witness was removed remains present with unavailable geometry. Reuse trials: at least one witness is assigned to several objects.}
\label{tab:missingwitness}
\begin{tabularx}{\linewidth}{@{}Xrrr@{}}\toprule
Policy & Count kept & Null orphan & Reuse trials\\\midrule
\RowsMissingWitness
\bottomrule\end{tabularx}\end{spacing}
\end{table}

Partial binding preserves the count and marks the orphan geometry unavailable in all trials. Nearest assignment redirects the orphan to an already used witness in 720 trials. The drop-unbound control removes one identity in every trial. Class-agnostic assignment reroutes the orphan in 96 trials and creates 136 cross-class links in total. Ungated same-class assignment reroutes it in 18 trials. This probe demonstrates the count/measurement-availability distinction under a controlled intervention; it does not establish the prevalence or detection accuracy of natural witness failures.

\subsection{Numerical verification}
The support-readout audit independently checks complete-component ownership, support gates, frozen semantic fields, normal calibration, strict decisions and task-macro AUROC/F1. The binding controls are refreshed by applying the same fixed reader to their previously saved object assignments, without a new neural or baseline run. The main policy reproduces every audited complete-support configuration and all 9,031 predictions exactly; each of the six policies retains 9,031 records under each reference mode. Controlled witness-removal trials and assignment traces are unchanged because they evaluate the binding interface before axis updates. Original manuscripts, master tables and prior experiment archives are preserved.

\section{Normal references and deployment}
For continuous operand $m_j$, let $q_{.01,j},q_{.99,j}$ be normal-fitting quantiles. The scale and development padding are
\begin{align}
b_j&=\max(q_{.99,j}-q_{.01,j},0.1|\operatorname{median}_{\mathcal T}m_j|,0.01),\\
p_j&=\max\left(0.25b_j,\max_{x\in\mathcal D}\{q_{.01,j}-m_j(x),m_j(x)-q_{.99,j},0\}\right).
\end{align}
The interval is $[L_j,U_j]=[q_{.01,j}-p_j,q_{.99,j}+p_j]$. Residual references use normal fitting observations; an unavailable reference key defaults to $\{0\}$. All required missing operands share one binary unavailable-evidence residual. Cached pixel, color, separator, cable and OCR observations are retained alongside reconstructed object-derived quantities.

Calibration selects the empirical 95th percentile with the \texttt{higher} convention. For $n$ sorted calibration scores, its one-based index is $1+\lceil.95(n-1)\rceil$. With ten normal calibration images per VID task, this selects the observed maximum. Decisions use the strict comparison $s>\tau$. The interval recipe, score normalization and calibration convention are shared across the internal configurations; references and thresholds correspond to each configuration's normal facts.

\par\medskip\begingroup
\centering\small\begin{spacing}{1.0}
\begin{paperalgorithm}{Normal-data configuration and SGDB deployment}\label{alg:pipeline}
\begin{tabularx}{\linewidth}{@{}r@{\quad}X@{}}
\textbf{Input} & Normal fitting, development and calibration roles; component targets; task requirements.\\[2pt]
1 & Train the visual frontend on component targets from normal fitting sources; select on normal development targets and freeze it.\\
2 & Train the predesignated semantic decoder on normal fitting parents and paired factual edits; select on normal development targets.\\
3 & Select the guarded screw readout on its annotated normal development subset.\\
4 & Select one association scheme on the 294 annotated original normal development parents.\\
5 & Fix the normal-development quality floor; apply the support-certified readout to bound normal objects, then fit intervals and residual references.\\
6 & Use independent normal calibration scores to freeze each task threshold.\\[2pt]
7 & \textbf{For an inspection image:} obtain decoded semantic objects and frozen geometric witnesses.\\
8 & Apply guarded screw geometry where eligible, then perform partial one-to-one association.\\
9 & Apply quality and complete-support checks; accept mask-axis updates or retain the initial axes. Rebuild operands and retain auxiliary appearance/OCR cues.\\
10 & Execute frozen task constraints and score normalization; compare the score strictly with its threshold.\\
11 & Record rule operands, residuals, normalized contributions and their source links; for an alarm, select the first sufficient descending prefix.\\
12 & Generate qualitative claims with the frozen language model, check their IDs and render numeric evidence from the common trace.\\
\textbf{Output} & Frozen score, image-level decision and supporting report.\\
\end{tabularx}
\end{paperalgorithm}
\end{spacing}
\par\endgroup

\section{Execution-grounded explanation}
\subsection{Trace schema and certificate selection}
Each trace record stores a rule key and evidence ID, observed value, executable requirement, residual, normalized score and source paths. Referenced objects retain their decoded indices, confidence, associated witness index, matching distance and available geometry. Length operands additionally retain the prior and candidate axes, support quality and completeness checks where a support-readout record exists. Cached pixel/OCR measurements retain their own source fields. A zero decoded count points to the count record rather than inventing an object region.

The certificate operates on the fixed normalized contributions. For a subset $C$, omitted scores are set to zero and $S(C)=\max_{j\in C}z_j+0.01\sum_{j\in C}z_j$, with $S(\varnothing)=0$. Sorting requires $O(J\log J)$ time and scanning the prefixes requires $O(J)$. For any subset $A$ of size $k$, its maximum is at most $z_{(1)}$ and its sum is at most $\sum_{i=1}^kz_{(i)}$. Therefore $S(A)\le S(C_k)$, proving minimum cardinality of the first prefix exceeding the threshold. Equal-score ties prefer available predicates over the aggregate unavailable-evidence flag and then use key order. This defines a sufficient score-space certificate; selected predicates may share underlying operands.

The full trace and score are preserved alongside the certificate. For non-alarms, the renderer retains unavailable operands and any raw deviations, including those normalized to zero. Otherwise it cites evaluated count predicates and the zero-score execution state. The language realization layer conditions on selected qualitative outcomes and evidence identifiers. Numeric values remain in the executor-owned evidence records. Image-level truth is attached only after generation for the qualitative review strip.

\subsection{Selection-policy comparison and checks}
All four policies use the same 66,297 rule records from 9,031 frozen images. The comparison cohort contains every one of the 5,725 original alarms. All-executed selection retains every rule; positive-score selection drops zero contributions; maximum-score selection retains one largest contribution; compact sufficient selection uses the first sufficient descending prefix. Mean size is an image-weighted average on that alarm cohort, and sufficiency uses each image's existing strict threshold comparison. No references, thresholds or detection predictions are changed.

\begin{table}[tbp]
\centering\small\begin{spacing}{1.05}
\caption{Per-category alarm counts and mean selected-rule counts. The last column counts alarms insufficiently explained by only the maximum-score rule. Every compact certificate is sufficient.}
\label{tab:tracecategories}
\begin{tabularx}{\linewidth}{@{}Xrrrrr@{}}\toprule
Category & Alarms & All & Positive & Compact & Max fail\\\midrule
\RowsExplanationCategories
\bottomrule\end{tabularx}\end{spacing}
\end{table}

The compact certificate retains 5,759 rule records, compared with 11,418 positive-score records and 40,767 executed records over the alarm cohort. Of the certificates, 5,708 contain one rule and 17 contain three. In the latter tape traces, each of the three selected contributions equals 4.044522437723423. One yields 4.084967662100657 and two yield 4.125412886477891, equal to the threshold; all three yield 4.165858110855125. The selected predicates are the green-tape count, red-tape count and measurement-unavailable flag.

The full reports retain missing evidence in 1,254 records, including 94 non-alarms; 191 non-alarm traces have raw deviations. Checks reconstruct every residual, normal-tail normalization and full score and reproduce all 9,031 decisions. Independent axis/center calculations reconstruct 3,928 cited geometric operands with maximum absolute error $5.684\times10^{-14}$. Certificate cardinality is also checked by exhaustive enumeration on 160 small synthetic score vectors, a joint-aggregation fixture and four deliberately corrupted report fields. These checks measure execution and provenance consistency.

\subsection{Frozen language realization}
The language input projects a trace record to its predicate name, operator, qualitative relation and evidence ID. Missing operands retain their names. It excludes raw image pixels, source filenames, image-level truth and detector-normalization values. The frozen model writes qualitative claims; the report renders observed numbers and requirements directly from the source records. The JSON response contains the unchanged decision and a list of claims with evidence IDs. Facts sharing one qualitative relation may be combined into a claim.

The contract checks the JSON schema, unchanged decision, exact selected-ID coverage, allowed citations and the absence of literal numeric quantities in generated text. It also checks relation grouping, explicit unavailable states and selected lexical conflicts. A rejected output is retained as a structured trace rather than accepted prose. These are mechanical preservation checks; the comparison measures contract adherence. All prompts, raw responses and rejection reasons are archived, including earlier prompt-adapter development runs.

The final renderer is evaluated on one SHA256-ranked source from each of the 55 deployment tasks, the six main-paper cases and three additional execution states: a joint alarm, a non-alarm with unavailable evidence and a non-alarm with raw deviations. Deduplication gives 63 English reports; seven case reports are also generated in Chinese. Selection does not consume image-level truth. The final batch runs on physical GPU 2 with the existing checkpoint, greedy decoding, batch size two and at most 640 new tokens per report. Detector facts, thresholds and decisions are unchanged. The model uses no image input and takes no optimization steps.

All 70 reports preserve the tested source contract (Table~\ref{tab:language}). Fifty deliberate corruptions of decisions, source IDs, claim coverage, numeric text and unsupported-cause text are rejected.

\begin{table}[!htb]
\centering\small\setlength{\tabcolsep}{5pt}\begin{spacing}{1.05}
\caption{Frozen language realization. Accepted denotes preservation of the defined source contract; mean claims measures report size. Numeric evidence is rendered by the executor.}
\label{tab:language}
\begin{tabularx}{\columnwidth}{@{}Xrrr@{}}\toprule
Language & Reports & Accepted & Mean claims\\\midrule
\RowsLanguage
\bottomrule\end{tabularx}\end{spacing}
\end{table}

\section{Comparison settings and condition results}
The external detection tables compare completed local deployments. LOCO ComAD uses DINO-S8 clustering with five clusters and CRF processing without component labels; its cross-scene deployment uses the archived shared component-map frontend. Local PUAD omits the external ImageNet penalty-image term. Cross-scene LogiCo uses its archived component frontend. LogicAD-FE-fast is a simplified local program/frontend deployment. Cross-scene SALAD and PUAD use category-level calibration, while SGDB uses deployment-task calibration. SGDB uses component annotations on the normal sources listed in Table~\ref{tab:roles}, factual edits with updated targets, and explicit task requirements.

Internal controls share the normal-source roles, component-supervision budgets, visual frontend, task requirements and score/calibration recipe. The plain-set and evidence-edit decoders each use 768 updates and 1,536 image exposures. Direct DFD and its initial bound readout share semantic predictions and differ in the source of geometric operands. Binding-policy controls share decoder outputs and the complete-support reader; geometric-readout controls share decoder outputs, initial witnesses and class masks. The comparison recomputes normal references and thresholds from each configuration's normal facts using the fixed recipe. The matched-reference diagnostic instead holds the initial intervals, references and thresholds fixed, measuring the decision changes due to axis selection alone.

\begin{table}[tbp]
\centering\small\begin{spacing}{1.05}
\caption{SGDB condition summaries across all ten VID scenarios. Values other than image counts are percentages. Missing is the task macro-average of unavailable-operand rates.}
\label{tab:conditions}
\begin{tabular}{lrrrrr}\toprule
Condition & Images & AUROC & F1 & Normal FPR & Missing\\\midrule
\RowsVidConditions
\bottomrule\end{tabular}\end{spacing}
\end{table}

\begin{spacing}{1.0}\small
\begin{longtable}{@{}p{23mm}p{45mm}p{49mm}p{29mm}@{}}
\caption{Information and implementation conditions of the completed deployments. NR: not established by the retained comparison protocol; the archive provides the actual image scores and thresholds. Internal mechanism controls share SGDB's component supervision and task requirements.}\label{tab:comparison-conditions}\\
\toprule Method & Normal data / component targets & Requirements / implementation & VID calibration\\\midrule\endfirsthead
\toprule Method & Normal data / component targets & Requirements / implementation & VID calibration\\\midrule\endhead
SGDB V2 & Fit/DEV/CAL roles; component targets in Table~\ref{tab:roles}. & Explicit requirements; learned semantics, bound geometry and support certification. & Task level.\\[3pt]
SALAD & Normal fitting; archived VID component-map input; LOCO annotation detail NR. & Local completed deployment; 70,000-step VID branch; no SGDB executor. & Category level.\\[3pt]
LogiCo & Normal fitting; archived VID component frontend; exact LOCO target lineage NR. & Local component adaptation; fixed 200-epoch VID fit; no SGDB executor. & Pooling NR.\\[3pt]
ComAD & LOCO normal RGB only, no component labels; VID shared component frontend. & LOCO DINO-S8, five clusters and CRF; VID component-map adapter. & Pooling NR.\\[3pt]
PUAD & LOCO normal RGB only; VID local shared component-map input. & 70,000 steps; external ImageNet penalty images omitted. & Category level.\\[3pt]
Qwen3-VL-8B & Archived prompts and normal score calibration; prompt-reference count NR. & Frozen image-language baseline; distinct from the explanation-side text model. & Pooling NR.\\[3pt]
LogicAD-FE-fast & Archived normal inputs and local frontend; target lineage NR. & Simplified local program/frontend deployment. & Pooling NR.\\
\bottomrule\end{longtable}\end{spacing}

The machine-readable condition matrix adds method-specific development selection, known annotation budgets, source-protocol links and unresolved adapter details. Detection tables describe these local deployments. Mechanism attribution uses the shared-supervision internal controls.

\section{All 50 VID task results}
The following tables report every retained primary method on all 50 deployment tasks. Each Mean row is an arithmetic task macro-average. The SGDB column refers to the fixed evidence-edit decoder with complete-support readout; the internal controls appear in the main-paper ablation. Best/second-best distinct displayed values are bold/underlined, and ties share marks.

\begin{spacing}{1.02}\fontsize{8.6}{10.5}\selectfont\setlength{\tabcolsep}{3pt}
\begin{longtable}{llrrrrrr>{\columncolor{OurBlue}}r}
\caption{Complete VID task AUROC (\%).}\label{tab:allvidauc}\\
\toprule Scenario & Condition & SALAD & LogiCo & ComAD & PUAD & Qwen & LogicAD & SGDB\\\midrule\endfirsthead
\multicolumn{9}{l}{Table~\ref{tab:allvidauc} continued}\\
\toprule Scenario & Condition & SALAD & LogiCo & ComAD & PUAD & Qwen & LogicAD & SGDB\\\midrule\endhead
\midrule\multicolumn{9}{r}{Continued on next page}\\\endfoot
\bottomrule\endlastfoot
\RowsVidAllAuroc
\end{longtable}
\begin{longtable}{llrrrrrr>{\columncolor{OurBlue}}r}
\caption{Complete VID fixed-threshold F1 (\%).}\label{tab:allvidf}\\
\toprule Scenario & Condition & SALAD & LogiCo & ComAD & PUAD & Qwen & LogicAD & SGDB\\\midrule\endfirsthead
\multicolumn{9}{l}{Table~\ref{tab:allvidf} continued}\\
\toprule Scenario & Condition & SALAD & LogiCo & ComAD & PUAD & Qwen & LogicAD & SGDB\\\midrule\endhead
\midrule\multicolumn{9}{r}{Continued on next page}\\\endfoot
\bottomrule\endlastfoot
\RowsVidAllF
\end{longtable}
\end{spacing}

\section{Paired grouped analysis and reproduction}

\subsection{Paired grouped uncertainty}
The paired analysis keeps the trained models, normal references and calibration thresholds fixed. It resamples groups with replacement within category and binary-label strata, using 10,000 replicates and seed 20261025. Each replicate uses the same group multiplicities for all compared methods. AUROC and fixed-threshold F1 are computed within each deployment task and then macro-averaged, matching the main tables. The reported intervals are the 2.5th and 97.5th percentiles of the paired differences in percentage points.

LOCO contributes 1,136 individual test-image groups. For VID, each scene, anomaly-path type and filename index forms a group of five capture conditions, giving 1,579 index groups for 7,895 images. This correspondence is an explicit resampling assumption; the retained dataset metadata do not verify physical-instance or source-video identity. Normal-role groups use the inherited manifest indices. Neither these group IDs nor exact image SHA-256 values overlap the fitting, development, calibration and test roles. This check does not establish independence between adjacent or visually similar frames.

\begin{table}[!htbp]\centering\small\begin{spacing}{1.0}
\caption{SGDB V2 minus each frozen control: point difference and paired grouped 95\% interval (percentage points). Each deployment retains its own archived normal references. The guarded-geometry comparison is secondary.}
\label{tab:grouped-ci}
\begin{tabular}{@{}llrr@{}}\toprule
Dataset & Control & $\Delta$ AUROC [95\% CI] & $\Delta$ F1 [95\% CI]\\\midrule
LOCO & Initial bound & +0.00 [+0.00, +0.00] & +0.00 [+0.00, +0.00] \\
LOCO & Uniform mask & +1.83 [+1.18, +2.51] & +0.98 [-0.07, +2.05] \\
LOCO & Quality only & +1.57 [+0.96, +2.22] & +0.70 [-0.28, +1.69] \\
LOCO & FMP + guarded geometry & +0.11 [+0.00, +0.29] & +0.98 [+0.06, +2.00] \\
VID-AD & Initial bound & +0.93 [+0.56, +1.32] & +0.97 [+0.57, +1.43] \\
VID-AD & Uniform mask & -0.02 [-0.08, +0.04] & -0.10 [-0.20, -0.02] \\
VID-AD & Quality only & +0.02 [+0.00, +0.06] & +0.00 [+0.00, +0.00] \\
VID-AD & FMP + guarded geometry & +2.22 [+1.49, +2.96] & +3.30 [+2.44, +4.19] \\
\bottomrule\end{tabular}\end{spacing}\end{table}

The intervals describe conditional variation within the evaluated test cohorts under the frozen configurations, rather than training-seed variation or independent blind validation. Each contrast has a marginal interval; the intervals are not adjusted for simultaneous multiple comparisons. The uniform-mask control has a small VID F1 advantage, while V2 protects LOCO AUROC against both unrestricted and quality-only axis replacement.

\begin{table}[!htbp]\centering\small\begin{spacing}{1.0}
\caption{Sensitivity analysis with the original bound deployment's intervals, residual references and thresholds held fixed for all readout configurations (percentage points). Same resampling draws as Table~\ref{tab:grouped-ci}.}
\label{tab:fixed-ci}
\begin{tabular}{@{}llrr@{}}\toprule
Dataset & Control & $\Delta$ AUROC [95\% CI] & $\Delta$ F1 [95\% CI]\\\midrule
LOCO & Initial bound & +0.00 [+0.00, +0.00] & +0.00 [+0.00, +0.00] \\
LOCO & Uniform mask & +0.68 [+0.33, +1.09] & +1.12 [+0.54, +1.71] \\
LOCO & Quality only & +0.15 [+0.02, +0.35] & +0.28 [-0.03, +0.64] \\
VID-AD & Initial bound & +0.93 [+0.56, +1.32] & +0.97 [+0.57, +1.43] \\
VID-AD & Uniform mask & -0.03 [-0.08, +0.02] & -0.05 [-0.13, +0.01] \\
VID-AD & Quality only & +0.02 [+0.00, +0.06] & +0.00 [+0.00, +0.00] \\
\bottomrule\end{tabular}\end{spacing}\end{table}

\begin{table}[!htbp]\centering\small\begin{spacing}{1.0}
\caption{Changes to V2 across all 9,031 images. R/G: repaired/regressed views. Group counts use LOCO image IDs and VID cross-condition filename indices; one group can contain both kinds of change. Complete unchanged and changed group ledgers are supplied.}
\label{tab:readout-transitions}
\begin{tabular}{@{}lrrrr@{}}\toprule
Control & R views & G views & R groups & G groups\\\midrule
Initial bound & 90 & 1 & 20 & 1 \\
Uniform mask & 28 & 27 & 28 & 23 \\
Quality only & 22 & 14 & 22 & 14 \\
\bottomrule\end{tabular}\end{spacing}\end{table}

\subsection{Reproduction material}
The CPU archive contains 162,558 saved predictions from 18 configurations, all image labels and source-index group IDs, frozen intervals/references/thresholds, and 12,666 final fact records. The metric script reproduces 990 method/task rows and checks all 715 archived manuscript rows. The scoring replay starts from facts, executes the archived scalar predicates and reconstructs score-space explanation certificates; it does not invoke a detector. Recorded language outputs have a separate contract replay. The archive also supplies exact group draws, per-category results, condition-specific false-positive rates, and source-index transition ledgers.

The code and frozen reproduction records are available in the public SGDB repository (\url{https://github.com/Qybc/SGDB}, release v1.0.0). The release includes the research implementations and 15 semantic-decoder checkpoints. Its model inventory records the visual-frontend assets outside this release; original-image inference also requires the benchmark data and OCR/model runtime. The supplied CPU entry points reconstruct the reported results from frozen outputs and facts. No new training, detector inference, baseline run or threshold fitting was performed for this analysis.

\section{Implementation consistency}
Consistency audits verify association, geometric provenance, normal references and strict decisions. They cover 1,176 development records, 25,332 initial facts, 18,062 initial predictions, 4,559 source-mask records and 36,124 complete-support predictions.

\section{Paired decisions and qualitative sources}
\subsection{Paired operating-point changes}
Table~\ref{tab:transitions} compares FMP plus guarded geometry with SGDB over the full image cohorts. Changes with identical raw residuals reflect configuration-specific normal references and calibration. Main-paper FPR values are task macro-averages.

\begin{table}[tbp]
\centering\small\begin{spacing}{1.05}
\caption{Paired decision transitions. Same residuals counts changed decisions with an identical raw rule-residual dictionary.}
\label{tab:transitions}
\begin{tabular}{lrrrrrr}\toprule
Dataset & \shortstack{Corrected\\FN} & \shortstack{New\\FN} & \shortstack{Corrected\\FP} & \shortstack{New\\FP} & Changes & \shortstack{Same\\residuals}\\\midrule
LOCO & 13 & 1 & 0 & 4 & 18 & 16\\
VID & 366 & 110 & 77 & 66 & 619 & 158\\
\bottomrule\end{tabular}\end{spacing}
\end{table}

\subsection{Qualitative source selection and replay}
The LOCO comparison uses the first two logical sources per category, followed by the first normal and third logical source. VID uses the first normal Original, anomalous Original and anomalous Blurry source per scenario. Detailed reports cover nine fixed sources, six in the main paper and three below. The joint-contribution inset adds the lexicographically first frozen tape trace requiring three contributions.

All reports use the audited execution traces. The six main reports use the frozen language model for qualitative claims, with numeric evidence rendered from source records; the three additional reports below retain the structured trace view. Ground truth is added only in the subsequent review strip.

Images are uniformly scaled and padded. Detail crops retain native coordinates and parent pixels. Boxes and axes show SGDB outputs; green/red strips indicate image-level correctness against the dataset label.

\begin{figure}[!htbp]
\centering\includegraphics[width=\textwidth,height=.88\textheight,keepaspectratio]{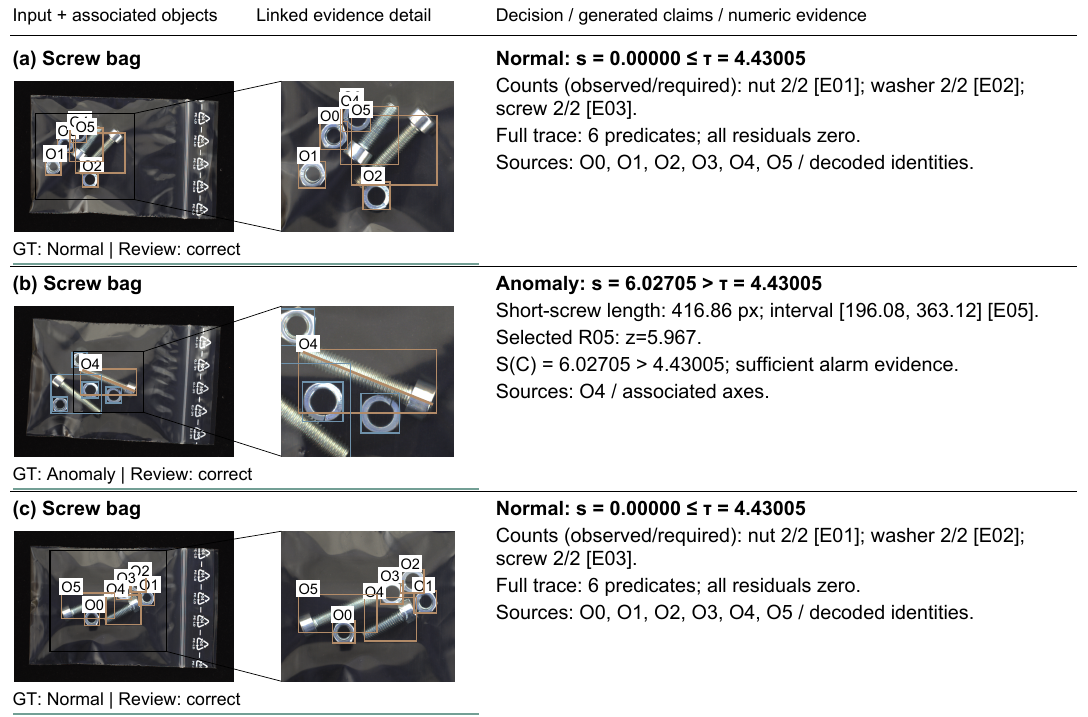}
\captionsetup{font={footnotesize,stretch=1.0}}
\caption{Additional SGDB reports on normal and anomalous screw bags. All decisions, lengths and interval references are from the current hybrid replay. The input and linked detail retain original dataset pixels.}
\label{fig:additional}
\end{figure}